\documentclass[journal]{IEEEtran}
\usepackage[dvips]{graphicx}
\usepackage{epsfig,multirow}
\usepackage{amsmath,amssymb,bm}
\usepackage{amsfonts,dsfont,color,bbm}
\usepackage{algorithm,algorithmic}
\usepackage[font=footnotesize,labelfont=bf]{caption}
\usepackage{subcaption,slashbox}
\usepackage{cite,epstopdf,epsf,epsfig}

\renewcommand{\vec}[1]{\mbox{\boldmath${#1}$}}
\newcommand{\norm}[1]{\left|\left|#1\right|\right|}

\newcommand{\Real}{{\mathbbm{R}}}
\newcommand{\Pre}{\mathcal{P}}
\newcommand{\R}{\mathcal{R}}
\newcommand{\Leaf}{\mathcal{L}}
\newcommand{\M}{\mathcal{M}}
\newcommand{\N}{\mathcal{N}}
\newcommand{\Sp}{\mathcal{S}}

\DeclareMathOperator*{\argmin}{arg\,min}

\newcommand{\sdt}{\left\{d_t \right\}_{t \geq 1}}
\newcommand{\svxt}{\left\{\vec{x}_t \right\}_{t \geq 1}}
\newcommand{\vhd}{\vec{\hat{d}}}
\newcommand{\vx}{\vec{x}}
\newcommand{\vw}{\vec{w}}

\newcommand{\vD}{\vec{D}}

\newcommand{\vI}{\vec{I}}
\newcommand{\vt}{{\vec{\theta}}}
\newcommand{\vv}{\vec{v}}
\newcommand{\hd}{\hat{\delta}}
\newcommand{\vmu}{\vec{\mu}}
\newcommand{\vsigma}{\vec{\Sigma}}
\newcommand{\vp}{\vec{\phi}}

\begin{document}
\renewcommand{\thepage}{}

\title{A Comprehensive Approach to Universal Piecewise Nonlinear Regression Based on Trees}

\author{N. Denizcan Vanli and Suleyman S. Kozat, \textit{Senior Member, IEEE}
\thanks{This work is supported in part by IBM Faculty Award and TUBITAK, Contract no: 112E161.}
\thanks{The authors are with the Department of Electrical and Electronics Engineering,
Bilkent University, Bilkent, Ankara 06800, Turkey, Tel: +90 (312)
290-2336, Fax: +90 (312) 290-1223, (e-mail: vanli@ee.bilkent.edu.tr,
kozat@ee.bilkent.edu.tr).} }

\maketitle
\begin{abstract}
In this paper, we investigate adaptive nonlinear regression and
introduce tree based piecewise linear regression algorithms that are
highly efficient and provide significantly improved performance with
guaranteed upper bounds in an individual sequence manner. We use
a tree notion in order to partition the space of regressors in a
nested structure. The introduced algorithms adapt not only their
regression functions but also the complete tree structure while
achieving the performance of the ``best'' linear mixture of a doubly
exponential number of partitions, with a computational complexity only
polynomial in the number of nodes of the tree.  While constructing these
algorithms, we also avoid using any artificial ``weighting'' of models
(with highly data dependent parameters) and, instead, directly
minimize the final regression error, which is the ultimate performance
goal. The introduced methods are generic such that they can readily
incorporate different tree construction methods such as random trees
in their framework and can use different regressor or partitioning
functions as demonstrated in the paper.
\end{abstract}
\begin{center}
\bfseries EDICS Category: ASP-ANAL, MLR-LEAR, MLR-APPL.
\end{center}

\section{Introduction}\label{sec:intro}
\PARstart{N}{onlinear} adaptive filtering and regression are extensively
investigated in the signal processing \cite{Singer3, Hero, drost, CTW, volterra, saf, fnf, add1, add2, add3, add4, add5, add6, add7, sp1, sp2, sp3, sp4, sp5}
and machine learning literatures \cite{Helmbold, ml1, ml2, ml3}, especially for
applications where linear modeling \cite{Singer1, Moon1} is
inadequate, hence, does not provide satisfactory results due to the
structural constraint on linearity. Although nonlinear approaches can
be more powerful than linear methods in modeling, they usually suffer
from overfitting, stability and convergence issues \cite{sayed_book,
  Singer3, Nascimento1, sayed2}, which considerably limit their
application to signal processing problems. These issues are especially
exacerbated in adaptive filtering due to the presence of feedback,
which is even hard to control for linear models \cite{sayed_book,
  Nascimento1, Garcia1}. Furthermore, for applications involving big
data, which require to process input vectors with considerably large
dimensions, nonlinear models are usually avoided due to unmanageable
computational complexity increase \cite{RPTrees}. To overcome these
difficulties, ``tree'' based nonlinear adaptive filters or regressors
are introduced as elegant alternatives to linear models since these
highly efficient methods retain the breadth of nonlinear models while
mitigating the overfitting and convergence issues
\cite{Hero,CTW,CTW2, RPTrees, ldf}.

\begin{figure}
  \centering
  \includegraphics[width=0.5\textwidth]{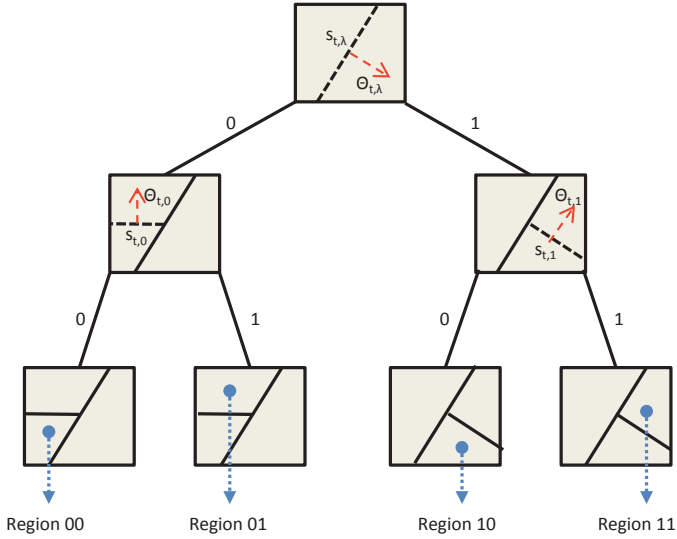}\\
  \caption{The partitioning of a two dimensional regressor space using
    a complete tree of depth-$2$ with hyperplanes for separation. The
    whole regressor space is first bisected by $s_{t,\lambda}$, which
    is defined by the hyperplane $\vt_{t,\lambda}$, where the
    region on the direction of $\vt_{t,\lambda}$ vector corresponds
    to the child with ``1'' label. We then
    continue to bisect children regions using $s_{t,0}$ and $s_{t,1}$,
    defined by $\vt_{t,0}$ and $\vt_{t,1}$,
    respectively.}\label{fig:Tree_Labels&Partitions}
\end{figure}

In its most basic form, a regression tree defines a hierarchical or
nested partitioning of the regressor space \cite{Hero}. As an example,
consider the binary tree in Fig. \ref{fig:Tree_Labels&Partitions},
which partitions a two dimensional regressor space.  On this tree,
each node represents a bisection of the regressor space, e.g., using
hyperplanes for separation, resulting a complete nested and disjoint
partition of the regressor space. After the nested partitioning is
defined, the structure of the regressors in each region can be chosen as desired, e.g.,
one can assign a linear regressor in each region yielding an overall
piecewise linear regressor. In this sense, tree based regression is a
natural nonlinear extension to linear modeling, in which the space of
regressors is partitioned into a union of disjoint regions where a
different regressor is trained. This nested architecture not only
provides an efficient and tractable structure, but also is shown to easily
accommodate to the intrinsic dimension of data, naturally alleviating
the overfitting issues \cite{RPTrees, KDTrees}.

Although nonlinear regressors using decision trees are powerful and
efficient tools for modeling, there exist several algorithmic
preferences and design choices that affect their performance
in real life applications \cite{CTW,CTW2,Hero}. Especially their adaptive
learning performance may greatly suffer if the algorithmic parameters
are not tuned carefully, which is particularly hard to accommodate for
applications involving nonstationary data exhibiting saturation
effects, threshold phenomena or chaotic behavior \cite{CTW}. In
particular, the success of the tree based regressors heavily depends
on the ``careful'' partitioning of the regressor space. Selection of a
good partition, including its depth and regions, from the hierarchy is
essential to balance the bias and variance of the regressor
\cite{RPTrees,KDTrees}. As an example, even for a uniform binary tree,
while increasing the depth of the tree improves the modeling power,
such an increase usually results in overfitting \cite{CTW}. There
exist numerous approaches that provide ``good'' partitioning of the
regressor space that are shown to yield satisfactory results on the
average under certain statistical assumptions on the data or on the
application \cite{RPTrees}.

We note that on the other extreme, there exist methods in adaptive
filtering and computational learning theory, which avoid such a direct
commitment to a particular partitioning but instead construct a
weighted average of all possible piecewise models defined on a tree
\cite{Takimoto1, Takimoto2,CTW}. Note that a full binary tree of
depth-$d$, as shown in Fig. \ref{fig:Tree_Labels&Partitions} for
$d=2$, with hard separation boundaries, defines a doubly exponential
number \cite{double} of complete partition of the regressor space (see
Fig. \ref{fig:Tree_Regressor_Space_Representation}). Each such
partitioning of the regressor space is represented by the collection
of the nodes of the full tree where each node is assigned to a
particular region of the regressor space. Any of these partitions can
be used to construct and then train a piecewise linear or nonlinear
regressor. Instead of fixing one of these partitions, one can run all
the models (or subtrees) in parallel and combine the final outputs
based on their performance. Such approaches are shown to mitigate the
bias variance trade off in a deterministic framework \cite{Takimoto1,
  Takimoto2,CTW}. However, these methods are naturally constraint to
work on a specific tree or partitionings, i.e., the tree is fixed and
cannot be adapted to the data, and the weighting among the models
usually have no theoretical justifications (although they may be
inspired from information theoretic considerations \cite{Willems}). As
an example, the ``universal weighting'' coefficients in
\cite{CTW,Singer1,Singer2,linder1,linder2} or the exponentially
weighted performance measure are defined based on algorithmic concerns
and provide universal bounds, however, do not minimize the final
regression error. In particular, the performance of these methods
highly depends on these weighting coefficients and algorithmic
parameters that should be tuned to the particular application for
successful operation \cite{CTW,CTW2}.

To this end, we provide a {\em comprehensive} solution to nonlinear
regression using decision trees. In this paper, we introduce
algorithms that are shown {\em i)} to be highly efficient {\em ii)} to provide
significantly improved performance over the state of the art
approaches in different applications {\em iii)} to have guaranteed
performance bounds without any statistical assumptions. Our algorithms
not only adapt the corresponding regressors in each region, but also
learn the corresponding region boundaries, as well as the ``best''
linear mixture of a doubly exponential number of partitions to
minimize the final estimation or regression error. We introduce
algorithms that are guaranteed to achieve the performance of the best
linear combination of a doubly exponential number of models with a
significantly reduced computational complexity. The introduced approaches significantly outperform
\cite{CTW,Singer1,Singer2} based on trees in different applications in
our examples, since we avoid any artificial weighting of models with
highly data dependent parameters and, instead, ``directly'' minimize
the final error, which is the ultimate performance goal. Our methods
are generic such that they can readily incorporate random projection
(RP) or $k$-d trees in their framework as commented in our
simulations, e.g., the RP trees can be used as the starting
partitioning to adaptively learn the tree, regressors and weighting to
minimize the final error as data progress.

In this paper, we first introduce an algorithm that asymptotically
achieves the performance of the ``best'' linear combination of a
doubly exponential number of different models that can be represented
by a depth-$d$ tree a with fixed regressor space partitioning with a
computational complexity only linear in the number of nodes of the
tree. We then provide a guaranteed upper bound on the performance of this algorithm and prove
that as the data length increases, this algorithm achieves the
performance of the ``best'' linear combination of a doubly exponential
number of models without any statistical assumptions. Furthermore,
even though we refrain from any statistical assumptions on the
underlying data, we also provide the mean squared performance of this
algorithm compared to the mean squared performance of the best linear
combination of the mixture. These methods are generic and truly
sequential such that they do not need any a priori information, e.g.,
upper bounds on the data \cite{Hero, CTW}, (such upper bounds does not
hold in general, e.g., for Gaussian data).  Although the combination
weights in \cite{CTW,Takimoto1, Takimoto2} are artificially constraint
to be positive and sum up to $1$ \cite{Garcia2}, we have no such
restrictions and directly adapt to the data without any
constraints. We then extend these results and provide the final
algorithm (with a slightly increased computational complexity), which
``adaptively'' learns also the corresponding regions of the tree to
minimize the final regression error. This approach learns {\em i)} the
``structure'' of the tree, {\em ii)} the regressors in each region, and {\em iii)}
the linear combination weights to merge all possible partitions, to
minimize the final regression error.  In this sense, this algorithm
can readily capture the salient characteristics of the underlying data
while avoiding bias to a particular model or structure.

In Section \ref{sec:hard_partitioning}, we first present an algorithm
with a fixed regressor space partitioning and present a guaranteed
upper bound on its performance. We then significantly reduce the
computational complexity of this algorithm using the tree
structure. In Section \ref{sec:soft_partitioning}, we extend these
results and present the final algorithm that adaptively learns the tree
structure, region boundaries, region regressors and combination
weights to minimize the final regression error.  We then demonstrate
the performance of our algorithms through simulations in Section
\ref{sec:Simulations}. We then finalize our paper with concluding
remarks.

\section{Problem Description}\label{sec:prob}
In this paper, all vectors are column vectors and denoted by boldface
lower case letters. Matrices are represented by boldface uppercase
letters. For a vector $\vec{u}$, $\norm{\vec{u}} =
\sqrt{\vec{u}^T\vec{u}}$ is the $\ell^2$-norm, where $\vec{u}^T$ is the
ordinary transpose. Here, $\vec{I}_k$ represents a $k\times k$ dimensional identity matrix.

We study sequential nonlinear regression, where we observe a desired
signal $\sdt$, $d_t \in \mathbbm{R}$, and regression vectors $\svxt$,
$\vx_t \in \mathbbm{R}^m$, such that we sequentially estimate $d_t$ by
\[
\hat{d}_t = f_t(\vx_t),
\]
and $f_t(\cdot)$ is an adaptive nonlinear regression function. At
each time $t$, the regression error is given by
\[
e_t = d_t -\hat{d}_t.
\]
Although there exist several different approaches to select the
corresponding nonlinear regression function, we particularly use
piecewise models such that the space of the regression vectors, i.e.,
$\vx_t \in \Real^m$, is adaptively partitioned using hyperplanes based
on a tree structure. We also use adaptive linear regressors in each
region. However, our framework can be generalized to any partitioning
of the regression space, i.e., not necessarily using hyperplanes, such
as using \cite{RPTrees}, or any regression function in each region,
i.e., not necessarily linear. Furthermore, both the region boundaries
as well as the regressors in each region are adaptive.

\subsection{A Specific Partition on a Tree}\label{sec:specific_partition}

\begin{figure}
  \centering
  \includegraphics[width=0.5\textwidth]{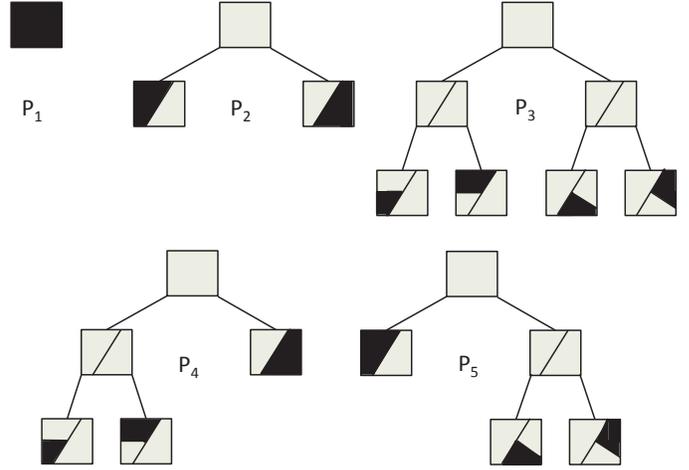}\\
  \caption{All different partitions of the regressor space that can be obtained using a depth-$2$ tree. Any of these partition can be used to construct a piecewise linear model, which can be adaptively trained to minimize the regression error. These partitions are based on the separation functions shown in  Fig. \ref{fig:Tree_Labels&Partitions}.}\label{fig:Tree_Regressor_Space_Representation}
\end{figure}

To clarify the framework, suppose the corresponding space of regressor
vectors is two dimensional, i.e., $\vx_t \in \mathbbm{R}^2$, and we
partition this regressor space using a depth-$2$ tree as in
Fig. \ref{fig:Tree_Labels&Partitions}.  A depth-$2$ tree is
represented by three separating functions $s_{t,\lambda}$, $s_{t,0}$
and $s_{t,1}$, which are defined using three hyperplanes with
direction vectors $\vt_{t,\lambda}$, $\vt_{t,0}$ and $\vt_{t,1}$,
respectively (See Fig.~\ref{fig:Tree_Labels&Partitions}). Due to the
tree structure, three separating hyperplanes generate only four
regions, where each region is assigned to a leaf on the tree given in
Fig. \ref{fig:Tree_Labels&Partitions} such that the partitioning is
defined in a hierarchical manner, i.e., $\vx_t$ is first processed by
$s_{t,\lambda}$ and then by $s_{t,i}$, $i=0,1$. A complete tree
defines a doubly exponential number, $O(2^{2^d})$, of subtrees each of
which can also be used to partition the space of past regressors. As
an example, a depth-$2$ tree defines 5 different subtrees or
partitions as shown in
Fig. \ref{fig:Tree_Regressor_Space_Representation}, where each of
these subtrees is constructed using the leaves and the nodes of the
original tree. Note that a node of the tree represents a region
which is the union of regions assigned to its left and right children
nodes \cite{Willems}.

The corresponding separating (indicator) functions can be hard, e.g.,
$s_{t}=1$ if the data falls into the region pointed by the
direction vector $\vt_{t}$, and $s_{t}=0$
otherwise. Without loss of generality, the regions pointed by the
direction vector $\vt_t$ are labeled as ``1'' regions on the tree in
Fig. \ref{fig:Tree_Labels&Partitions}. The separating functions can
also be soft. As an example, we use the logistic regression classifier \cite{logistic}
\begin{equation}\label{eq:separator_not_abused}
    s_t = \frac{1}{1+e^{\vx_t^T \vt_t+b_t}}
\end{equation}
as the soft separating function, where $\vt_t$ is the direction vector
and $b_t$ is the offset, describing a hyperplane in the
$m$-dimensional regressor space. With an abuse of notation we combine
the direction vector $\vt_t$ with the offset parameter $b_t$ and
denote it by $\vt_t = [\vec{\theta}_t;b_t]$. Then the separator
function in \eqref{eq:separator_not_abused} can be rewritten as
\begin{equation}\label{eq:separator}
    s_t = \frac{1}{1+e^{\vx_t^T \vt_t}},
\end{equation}
where $\vx_t = [\vx_t;1]$. One can easily use other differentiable
soft separating functions in this setup in a straightforward manner as
remarked later in the paper.

To each region, we assign a regression function to generate an
estimate of $d_t$. For a depth-$2$ (or a depth-$d$) tree, there are
$7$ (or $2^{d+1}-1$) nodes (including the leaves) and $7$ (or
$2^{d+1}-1$) regions corresponding to these nodes, where the
combination of these nodes or regions form a complete partition. In
this paper, we assign linear regressors to each region. For instance
consider the third model in
Fig. \ref{fig:Tree_Regressor_Space_Representation}, i.e., $P_3$, where
this partition is the union of 4 regions each corresponding to a leaf
of the original complete tree in Fig.~\ref{fig:Tree_Labels&Partitions}, labeled
as $00$, $01$, $10$, and $11$. The $P_3$ defines a complete
partitioning of the regressor space, hence can be used to construct a
piecewise linear regressor.  At each region, say the $00$th region, we
generate the estimate
\begin{equation}\label{eq:node_regressor}
\hat{d}_{t,00} = \vx_t^T\vec{v}_{t,00},
\end{equation}
where $\vec{v}_{t,00} \in \mathbbm{R}^m$ is the linear regressor
vector assigned to region $00$.
Considering the hierarchical structure of the tree and having
calculated the region estimates, the final estimate of $P_3$ is given
by
\begin{align}\label{eq:final_estimate}
\hat{d}_t =& s_{t,\lambda} s_{t,0}  \hat{d}_{t,00} + s_{t,\lambda} (1-s_{t,0})  \hat{d}_{t,01} \nonumber\\
           & + (1-s_{t,\lambda}) s_{t,1}  \hat{d}_{t,10} + (1-s_{t,\lambda}) (1-s_{t,1})  \hat{d}_{t,11},
\end{align}
for any $\vx_t$. We emphasize that any $P_i$, $i=1,\ldots,5$ can be used in a similar fashion to construct a piecewise linear regressor.

Continuing with the specific partition $P_3$, we adaptively train the
region boundaries and regressors to minimize the final regression
error. As an example, if we use a stochastic gradient descent algorithm
\cite{Hazan, Eweda, Garcia2, Garcia3}, we update the
regressor of the node ``00'' as
\begin{align}
    \vv_{t+1,00} &= \vv_{t,00} - \frac{1}{2} \mu_t \nabla e^2_t \nonumber\\
                 &= \vv_{t,00} + \mu_t e_t s_{t,\lambda} s_{t,0} \vx_t, \nonumber
\end{align}
where $\mu_t$ is the step size to update the region regressors. Similarly, region regressors can be updated for all
regions $r=00,01,10,11$.  Separator functions can also be
trained using the same approach, e.g., the separating function of the
node $0$,  $s_{t,0}$, can be updated as
\begin{align}
    \vec{\theta}_{t+1,0} &= \vec{\theta}_{t,0} - \frac{1}{2} \eta_t \nabla e^2_t \nonumber\\
                         &= \vec{\theta}_{t,0} + \eta_t e_t \big{(} s_{t,\lambda} \hat{d}_{t,00} - s_{t,\lambda} \hat{d}_{t,01} \big{)} \frac{\partial s_{t,0}}{\partial \vt_{t,0}}, \nonumber
\end{align}
where $\eta_t$ is the step size to update the separator functions and
\begin{equation}\label{eq:separator_der}
    \frac{\partial s_{t,0}}{\partial \vt_{t,0}} = \frac{-\vx_t e^{\vx_t^T \vec{\theta}_{t,0}}}{\big{(} 1+e^{\vx_t^T \vec{\theta}_{t,0}} \big{)}^2},
\end{equation}
according to the separator function in \eqref{eq:separator}. Other separating functions (different than the logistic regressor classifier) can also be trained
in a similar fashion by simply calculating the gradient with respect
to the extended direction vector and plugging in
\eqref{eq:separator_der}.

Until now a specific partition, i.e., $P_3$, is used to construct a
piecewise linear regressor, although the tree can represent $P_i$,
$i=1,\ldots,5$. However, since the data structure is unknown, one may
not prefer a particular model \cite{Takimoto1, Takimoto2,CTW}, i.e.,
there may not be a specific best model or the best model can change in
time. As an example, the simpler models, e.g., $P_1$, may perform
better while there is not sufficient data at the start of training and
the finer models, e.g., $P_3$, can recover through the learning process.
Hence, we hypothetically construct all doubly exponential number of
piecewise linear regressors corresponding to all partitions (see
Fig. \ref{fig:Tree_Regressor_Space_Representation}) and then calculate
an adaptive linear combination of the outputs of all, while these
algorithms learn the region boundaries as well as the regressors in
each region.

In Section \ref{sec:hard_partitioning}, we first consider the scenario
in which the regressor space is partitioned using hard separator
functions and  combine $O(2^{2^d})$ different models
for a depth-$d$ tree with a computational complexity $O(d2^d)$. In
Section \ref{sec:soft_partitioning}, we partition the regressor space
with soft separator functions and adaptively update the region
boundaries to achieve the best partitioning of the $m$-dimensional
regressor space with a computational complexity $O(m4^d)$.

%As shown in Table 1, for a depth-$d$ tree, we require $O(2^d\times m)$
%computations to adapt and calculate the estimate at each region and
%$O(2^d)$ computations to adapt the region boundaries and calculate the
%final estimate in \eqref{eq:regress_first}.

%partitioned using a tree of a fixed depth, say depth $d$, as shown in
%Fig. \ref{fig:basic_tree}.

\section{Regressor Space Partitioning via Hard Separator Functions}\label{sec:hard_partitioning}
In this section, we consider the regression problem in which the
sequential regressors (as described in Section
\ref{sec:specific_partition}) for all partitions in the doubly
exponential tree class are combined when hard separation functions are
used, i.e., $s_t \in \{0,1\}$. In this section, the hard boundaries
are not trained, however, both the regressors of each region and the
combination parameters to merge the outputs of all partitions are
trained. To partition the regressor space, we first construct a tree
with an arbitrary depth, say a tree of depth-$d$, and denote the
number of different models of this class by $\beta_d \approx
(1.5)^{2^d}$, e.g., one can use RP trees as the starting tree
\cite{RPTrees}. While the $k$th model (i.e., $P_k$ partition)
generates the regression output $\hat{d}_t^{(k)}$ at time $t$ for all
$k=1,\dots,\beta_d$, we linearly combine these estimates using the
weighting vector $\vw_t \triangleq
[w_t^{(1)},\dots,w_t^{(\beta_d)}]^T$ such that the final estimate of
our algorithm at time $t$ is given as
\begin{align}\label{eq:weighted_combination}
    \hat{d}_t & \triangleq \sum_{k=1}^{\beta_d} w_t^{(k)} \hat{d}_t^{(k)} \nonumber\\
              & = \vw_t^T \vhd_t,
\end{align}
where $\vhd_t \triangleq
[\hat{d}_t^{(1)},\dots,\hat{d}_t^{(\beta_d)}]^T$. The regression error
at time $t$ is calculated as
\begin{align}
    e_t(\vw_t) & \triangleq d_t - \hat{d}_t \nonumber\\
               & = d_t - \vw_t^T \vhd_t. \nonumber
\end{align}
For $\beta_d$ different models that are embedded within a depth-$d$
tree, we introduce an algorithm (given in Algorithm \ref{alg:Theorem1})
that asymptotically achieves the same cumulative squared regression
error as the optimal linear combination of these models without any
statistical assumptions.  This algorithm is constructed in the proof
of the following theorem and the computational complexity of the
algorithm is only linear in the number of the nodes of the tree.

{\bf Theorem 1: }{\em Let $\sdt$ and $\svxt$ be arbitrary, bounded, and real-valued sequences. The algorithm $\hat{d}_t$ given in Algorithm \ref{alg:Theorem1} when applied to these data sequence yields
\begin{equation}\label{eq:Theorem1}
   \sum_{t=1}^n \big{(} d_t - \hat{d}_t \big{)}^2 - \min_{\vw \in \Real^{\beta_d}} \sum_{t=1}^n \big{(} d_t - \vw^T\vhd_t \big{)}^2 \leq O \big{(} \ln(n) \big{)},
\end{equation}
for all $n$, when $e_t^2(\vw)$ is strongly convex $\forall t$, where
$\vhd_t=[\hat{d}_t^{(1)},\dots,\hat{d}_t^{(\beta_d)}]^T$, and
$\hat{d}_t^{(k)}$ are the estimates of $d_t$ at time $t$ for
$k=1,\dots,\beta_d$.}

This theorem implies that our algorithm (given in Algorithm
\ref{alg:Theorem1}), asymptotically achieves the performance of the
best combination of the outputs of $O(2^{2^d})$ different models that
can be represented using a depth-$d$ tree with a computational
complexity $O(d2^d)$. Note that as given in Algorithm \ref{alg:Theorem1}, no a priori information, e.g., upper bounds, on the data is used to construct the algorithm. Furthermore, the algorithm can use different regressors, e.g.,
\cite{CTW}, or regions seperation functions, e.g.,  \cite{RPTrees}, to define the tree.

Assuming that the constituent partition regressors converge to
stationary distributions, such as for Gaussian regressors, and under
widely used separation assumptions \cite{sayed_book,kozat} such that
the expectation of $\hat{d}_t^{(k)}$, $k=1,\dots,\beta_d$, and $\vw_t$
are separable, we have the following theorem.

{\bf Theorem 2: }{\em Assuming that the partition regressors, i.e.,
  $\hat{d}_t^{(k)}$, $k=1,\ldots,\beta_d$, and $d_t$ converge to zero mean
  stationary distributions, we have
\[
\lim_{t \to \infty} E[e^2_t] = J^* + \frac{\mu \, J^* \, \mathrm{tr}( \vD )}{2 - \mu \, \mathrm{tr}( \vD )},
\]
where $\mu$ is the learning rate of the stochastic gradient update,
\[
J^*  \triangleq \min_{\vw \in \Real^{\beta_d}} \lim_{t \to \infty} E[ (d_t - \vw^T \vhd_t)^2],
\]
and
\[
\vD \triangleq \lim_{t \to \infty} E \left[ \vhd_t \vhd_t^T \right],
\]
for the algorithm $\hat{d}_t$ (given in Algorithm \ref{alg:Theorem1}).}

Theorem 2 directly follows Chapter 6 of \cite{sayed_book} since we use a
stochastic gradient algorithm to merge the partition regressors
\cite{sayed_book, kozat}. Hence, the introduced algorithm may also
achieve the mean square error performance of the best linear
combination of the constituent piecewise regressors if $\mu$ is
selected carefully.

\subsection{Proof of Theorem 1 and Construction of Algorithm 1}
To construct the final algorithm, we first introduce a ``direct''
algorithm which achieves the corresponding bound in Theorem 1. This
direct algorithm has a computational complexity $O(2^{2^d})$ since one
needs to calculate the correlation information of $O(2^{2^d})$ models
to achieve the performance of the best linear combination.  We then
introduce a specific labeling technique and using the properties of
tree structure, construct an algorithm to obtain the same upper bound
as the ``direct'' algorithm, yet with a significantly smaller
computational complexity, i.e., $O(d2^d)$.

For a depth-$d$ tree, suppose $\hat{d}_{t}^{(k)}$, $k=1,\dots,\beta_d$,
are obtained as described in Section
\ref{sec:specific_partition}. To achieve the upper bound
in \eqref{eq:Theorem1}, we use the stochastic gradient
descent approach and update the combination weights as
\begin{align}\label{eq:GradientDescent}
  \vw_{t+1} &= \vw_t - \frac{1}{2} \mu_t \nabla e_t^2(\vw_t) \nonumber\\
            &= \vw_t + \mu_t e_t \vhd_t,
\end{align}
where $\mu_t$ is the step-size parameter (or the learning rate) of the
gradient descent algorithm. We first derive an upper bound on the
sequential learning regret $R_n$, which is defined as
\[
R_n \triangleq \sum_{t=1}^n e_t^2(\vw_t) - \sum_{t=1}^n e_t^2(\vw^*_n),
\]
where $\vw^*_n$ is the optimal weight vector over $n$, i.e.,
\begin{align}
\vw^*_n  \triangleq \argmin_{\vw \in \Real^{\beta_d}} \sum_{t=1}^n e_t^2(\vw). \nonumber
\end{align}
Following \cite{Hazan}, using Taylor series approximation, for some point $\vec{z}_t$ on the line segment connecting $\vw_t$ to $\vw^*_n$, we have
\begin{align}\label{eq:first}
    e_t^2(\vw^*_n) = ~& e_t^2(\vw_t) + \left(\nabla e_t^2(\vw_t)\right)^T (\vw^*_n - \vw_t) \nonumber\\
                      & + \frac{1}{2} (\vw^*_n-\vw_t)^T \nabla^2 e_t^2(\vec{z}_t) (\vw^*_n - \vw_t).
\end{align}
According to the update rule in \eqref{eq:GradientDescent}, at each iteration the update on weights are performed as $\vw_{t+1} = \vw_t - \frac{\mu_t}{2} \nabla e_t^2(\vw_t)$. Hence, we have
\begin{align}
  \norm{\vw_{t+1}-\vw^*_n}^2 &= \norm{\vw_t - \frac{\mu_t}{2} \nabla e_t^2(\vw_t) - \vw^*_n}^2 \nonumber\\
        & = \norm{\vw_t - \vw^*_n}^2 \hspace{-0.1cm} - \hspace{-0.1cm} \mu_t \left(\nabla e_t^2(\vw_t)\right)^T \hspace{-0.1cm} (\vw_t - \vw^*_n) \nonumber\\
         & \hspace{0.5cm} + \frac{\mu_t^2}{4} \norm{\nabla e_t^2(\vw_t)}^2. \nonumber
\end{align}
Then we obtain
\begin{align}
\left(\nabla e_t^2(\vw_t)\right)^T (\vw_t - \vw^*_n) & = \frac{\norm{\vw_t - \vw^*_n}^2 - \norm{\vw_{t+1}-\vw^*_n}^2}{\mu_t} \nonumber\\
    & \hspace{0.5cm} + \mu_t \frac{\norm{\nabla e_t^2(\vw_t)}^2}{4}.
\end{align}
Under the mild assumptions that $\norm{\nabla e_t^2(\vw_t)}^2 \leq A^2$ for some $A>0$ and $e_t^2(\vw^*_n)$ is $\lambda$-strong convex for some $\lambda>0$ \cite{Hazan}, we achieve the following upper bound
\begin{align}\label{eq:SGD_Regret_t}
    e_t^2(\vw_t) - e_t^2(\vw^*_n) & \leq \frac{\norm{\vw_t - \vw^*_n}^2 - \norm{\vw_{t+1}-\vw^*_n}^2}{\mu_t} \nonumber\\
     & \hspace{0.5cm} - \frac{\lambda}{2} \norm{\vw_t - \vw^*_n}^2 + \mu_t \frac{A^2}{4}.
\end{align}
By selecting $\mu_t = 2/(\lambda t)$ and summing up the regret terms in \eqref{eq:SGD_Regret_t}, we get
\begin{align}
  R_n &= \sum_{t=1}^n \left\{ e_t^2(\vw_t) - e_t^2(\vw^*_n) \right\} \nonumber\\
      & \leq \sum_{t=1}^n \norm{\vw_t - \vw^*_n}^2 \bigg{(} \frac{1}{\mu_t} - \frac{1}{\mu_{t-1}} - \frac{\lambda}{2} \bigg{)} + \frac{A^2}{4} \sum_{t=1}^n \mu_t \nonumber\\
      & = \frac{A^2}{4} \sum_{t=1}^n \frac{2}{\lambda t} \nonumber\\
      & \leq \frac{A^2}{2\lambda} \left(1 + \log(n) \right). \nonumber
\end{align}
Note that \eqref{eq:GradientDescent} achieves the performance
of the best linear combination of $O(2^{2^d})$ piecewise linear models
that are defined by the tree. However,
in this form \eqref{eq:GradientDescent} requires a computational
complexity of $O(2^{2^d})$ since the vector $\vw_t$ has a size of
$O(2^{2^d})$. We next illustrate an algorithm that performs the same
adaptation in \eqref{eq:GradientDescent} with a complexity of
$O(d2^d)$.

We next introduce a labeling for the tree nodes following
\cite{Willems}. The root node is labeled with an empty binary string
$\lambda$ and assuming that a node has a label $p$, where $p$ is a
binary string, we label its upper and lower children as $p1$ and $p0$,
respectively. Here we emphasize that a string can only take its
letters from the binary alphabet $\{0,1\}$, where $0$ refers to the
lower child, and $1$ refers to the upper child of a node. We also
introduce another concept, i.e., the definition of the prefix of a
string. We say that a string $p' = q'_{1} \dots q'_{l'}$ is a prefix
to string $p = q_1 \dots q_l$ if $l' \leq l$ and $q'_i = q_i$ for all
$i=1,\dots,l'$, and the empty string $\lambda$ is a prefix to all
strings. Let $\Pre(p)$ represent all prefixes to the string $p$, i.e.,
$\Pre(p) \triangleq \{ \nu_1,\dots,\nu_{l+1} \}$, where $l \triangleq
l(p)$ is the length of the string $p$, $\nu_i$ is the string with
$l(\nu_i) = i-1$, and $\nu_1 = \lambda$ is the empty string, such that
the first $i-1$ letters of the string $p$ forms the string $\nu_i$ for
$i=1,\dots,l+1$.

We then observe that the final estimate of any model can be found as the combination of the regressors of its leaf nodes. According to the region $\vx_t$ has fallen, the final estimate will be calculated with the separator functions. As an example, for the second model in Fig. \ref{fig:Tree_Regressor_Space_Representation} (i.e., $P_2$ partition), say $\vx_t \in \R_{00}$, and hard separator functions are used. Then the final estimate of this model will be given as $\hat{d}_t^{(2)} = \hat{d}_{t,0}$. For any separator function, the final estimate of the desired data $d_t$ at time $t$ of the $k$th model, i.e., $\hat{d}_t^{(k)}$ can be obtained according to the hierarchical structure of the tree as the sum of regressors of its leaf nodes, each of which are scaled by the values of the separator functions of the nodes between the leaf node and the root node. Hence, we can compactly write the final estimate of the $k$th model at time $t$ as
\begin{equation}\label{eq:General_Final_Estimate}
  \hat{d}_t^{(k)} = \sum_{p \in \M_k} \bigg{(} \hat{d}_{t,p} \prod_{i=1}^{l(p)} s_{t,\nu_i}^{q_i} \bigg{)},
\end{equation}
where $\M_k$ is the set of all leaf nodes in the $k$th model, $\hat{d}_{t,p}$ is the regressor of the node $p$, $l(p)$ is the length of the string $p$, $\nu_i \in \Pre(p)$ is the prefix to string $p$ with length $i-1$, $q_i$ is the $i$th letter of the string $p$, i.e., $\nu_{i+1} =\nu_i q_i$, and finally $s_{t,\nu_i}^{q_i}$ denotes the separator function at node $\nu_i$ such that
\[
  s_{t,\nu_i}^{q_i} \triangleq
    \begin{cases}
        s_{t,\nu_i}   ,& \text{if } q_i = 0 \\
        1-s_{t,\nu_i} ,& \text{otherwise}
    \end{cases}
\]
with $s_{t,\nu_i}$ defined as in \eqref{eq:separator}. We emphasize that we dropped $p$-dependency of $q_i$ and $\nu_i$ to simplify notation.

As an example, if we consider the third model $P_3$ in Fig. \ref{fig:Tree_Regressor_Space_Representation} as the $k$th model (i.e., $k=3$), where $\M_k = \{00,01,10,11\}$, then we can calculate the final estimate of that model as follows
\begin{align}\label{eq:General_Final_Estimate_Example}
  \hat{d}_t^{(k)} &= \sum_{p \in \M_k} \bigg{(} \hat{d}_{t,p} \prod_{i=1}^{l(p)} s_{t,\nu_i}^{q_i} \bigg{)} \nonumber\\
                  &= \hat{d}_{t,00} s_{t,0}^{0} s_{t,\lambda}^{0} + \hat{d}_{t,01} s_{t,0}^{1} s_{t,\lambda}^{0} \nonumber\\
                  &\hspace{0.2cm} + \hat{d}_{t,10} s_{t,1}^{0} s_{t,\lambda}^{1} + \hat{d}_{t,11} s_{t,1}^{1} s_{t,\lambda}^{1} \nonumber\\
                  &= \hat{d}_{t,00} s_{t,0} s_{t,\lambda} + \hat{d}_{t,01} \big{(} 1-s_{t,0} \big{)} s_{t,\lambda} \nonumber\\
                  &\hspace{0.2cm} + \hat{d}_{t,10} s_{t,1} \big{(} 1-s_{t,\lambda} \big{)} + \hat{d}_{t,11} \big{(} 1-s_{t,1} \big{)} \big{(} 1-s_{t,\lambda} \big{)}.
\end{align}
Note that \eqref{eq:final_estimate} and \eqref{eq:General_Final_Estimate_Example} are the same special cases of \eqref{eq:General_Final_Estimate}.

We next denote the product terms in \eqref{eq:General_Final_Estimate} as follows
\begin{equation}\label{eq:Delta_Definition}
  \hd_{t,p} \triangleq \hat{d}_{t,p} \prod_{i=1}^{l(p)} s_{t,\nu_i}^{q_i},
\end{equation}
to simplify the notation. Here, $\hd_{t,p}$ can be viewed as the estimate of the node (i.e., region) $p$ given that $\vx_t \in \R_{p'}$ for some $p' \in \Leaf_d$, where $\Leaf_d$ denotes all leaf nodes of the depth-$d$ tree class, i.e., $\Leaf_d \triangleq \{ p  :  l(p)=d \}$. Then \eqref{eq:General_Final_Estimate} can be rewritten as follows
\[
\hat{d}_t^{(k)} = \sum_{p \in \M_k} \hd_{t,p}.
\]

Since we now have a compact form to represent the tree and the outputs of each partition, we next introduce a method to calculate the combination weights of $O(2^{2^d})$ piecewise regressor outputs in a simplified manner.

To this end, we assign a particular linear weight to each node. We denote the weight of node $p$ at time $t$  as $w_{t,p}$ and then we define the weight of the $k$th model as the sum of weights of its leaf nodes, i.e.,
\[
w_t^{(k)} = \sum_{p \in \M_k} w_{t,p},
\]
for all $k = 1,\dots,\beta_d$. Since the weight of each model, say model $k$, is recursively updated as
\[
w_{t+1}^{(k)} = w_t^{(k)} + \mu_t e_t \hat{d}_t^{(k)},
\]
we achieve the following recursive update on the node weights
\begin{equation}\label{eq:node_weight}
  w_{t+1,p} \triangleq w_{t,p} + \mu_t e_t \hd_{t,p},
\end{equation}
where $\hd_{t,p}$ is defined as in \eqref{eq:Delta_Definition}.

This result implies that instead of managing $O(2^{2^d})$ memory locations, and making $O(2^{2^d})$ calculations, only keeping track of the weights of every node is sufficient, and the number of nodes in a depth-$d$ model is $|\N_d| = 2^{d+1}-1$, where $\N_d$ denotes the set of all nodes in a depth-$d$ tree. As an example, for $d=2$ we obtain $\N_d = \{\lambda,0,1,00,01,10,11\}$. Therefore we can reduce the storage and computational complexity from $O(2^{2^d})$ to $O(2^d)$ by performing the update in \eqref{eq:node_weight} for all $p \in \N_d$. We then continue the discussion with the update of weights performed at each time $t$ when hard separator functions are used.

Without loss of generality assume that at time $t$, the regression vector $\vx_t$ has fallen into the region $\R_{p'}$ specified by the node $p' \in \Leaf_d$. Consider the node regressor defined in \eqref{eq:Delta_Definition} for some node $p \in \N_d$. Since we are using hard separator functions, we obtain
\[
  \hd_{t,p} =
    \begin{cases}
        \hat{d}_{t,p} ,& \text{if } p \in \Pre(p') \\
        0 ,& \text{otherwise}
    \end{cases},
\]
where $\Pre(p')$ represents all prefixes to the string $p'$, i.e., $\Pre(p') = \{\nu'_1,\dots,\nu'_{d+1}\}$. Then at each time $t$ we only update the weights of the nodes $p \in \Pre(p')$, hence we only make $|\Pre(p')| = d+1$ updates since the hard separation functions are used for partitioning of the regressor space.

Before stating the algorithm that combines these node weights as well as node estimates, and generates the same final estimate as in \eqref{eq:weighted_combination} with a significantly reduced computational complexity, we observe that for a node $p \in \N_d$ with length $l(p) \geq 1$, there exist a total of
\[
\gamma_d\big{(}l(p)\big{)} \triangleq \prod_{j=1}^{l(p)}\beta_{d-j}
\]
different models in which the node $p \in \N_d$ is a leaf node of that model, where $\beta_0 = 1$ and $\beta_{j+1} = \beta_j^2+1$ for all $j\geq1$. For $l(p)=0$ case, i.e., for $p = \lambda$, one can clearly observe that there exists only one model having $\lambda$ as the leaf node, i.e., the model having no partitions, therefore $\gamma_d(0) = 1$.

Having stated how to store all estimates and weights in $O(2^d)$ memory locations, and perform the updates at each iteration, we now introduce an algorithm to combine them in order to obtain the final estimate of our algorithm, i.e., $\hat{d}_t = \vw_t^T \vhd_t$. We emphasize that the sizes of the vectors $\vw_t$ and $\vhd_t$ are $O(2^{2^d})$, which forces us to make $O(2^{2^d})$ computations. We however introduce an algorithm with a complexity of $O(d2^d)$ that is able to achieve the exact same result.

%\begin{table}
%  \centering
    \begin{algorithm}
        \caption{Decision Fixed Tree (DFT) Regressor}\label{alg:Theorem1}
        \begin{algorithmic}[1]
            \FOR{$t=1$ \TO $n$}
                \STATE $p' \Leftarrow p \in \Leaf_d : \vx_t \in \R_p$
                \STATE $\hat{d}_t \Leftarrow 0$
                \FORALL{$\nu_j' \in \Pre(p')$}
                    \STATE $\hat{d}_{t,\nu_j'} \Leftarrow \vv_{t,\nu_j'}^T \vx_t$
                    \STATE $\kappa_{t,\nu_j'} \Leftarrow \gamma_d\big{(}l(\nu_j)\big{)} w_{t,\nu_j'}$
                    \FORALL{$p \in \N_d-(\Pre(p') \cup \Sp_d(p'))$}
                        \STATE $\bar{p} \Leftarrow \acute{p} \in \Pre(p) \cap \Pre(p') : l(\acute{p}) = \left| \Pre(p) \cap \Pre(p') \right|-1$
                        \STATE $\kappa_{t,\nu_j'} \Leftarrow \kappa_{t,\nu_j'} + \frac{ \gamma_d\big{(}l(\nu_j')\big{)} \gamma_{d-l(\bar{p})-1}\big{(}l(p)-l(\bar{p})-1\big{)} }{ \beta_{d-l(\bar{p})-1} } w_{t,p}$
                    \ENDFOR
                \STATE $\hat{d}_t \Leftarrow \hat{d}_t + \kappa_{t,\nu_j'} \hat{d}_{t,\nu_j'}$
                \ENDFOR
                \STATE $e_t \Leftarrow d_t - \hat{d}_t$
                \FORALL{$\nu_j' \in \Pre(p')$}
                    \STATE $\vv_{t+1,\nu_j'} \Leftarrow \vv_{t,\nu_j'} + \mu_t e_t \vx_t$
                    \STATE $w_{t+1,\nu_j'} \Leftarrow w_{t,\nu_j'} + \mu_t e_t \hat{d}_{t,\nu_j'}$
                \ENDFOR
            \ENDFOR
        \end{algorithmic}
    \end{algorithm}
%  \caption{Algorithm I: Online Piecewise Linear Regressor for Hard Partitioning of the Regressor Space}\label{alg:Theorem1}
%\end{table}

For a depth-$d$ tree, at time $t$ say $\vx_t \in \R_{p'}$ for a node $p' \in \Leaf_d$. Then the final estimate of our algorithm is found by
\begin{align}\label{eq:final_estimate_doubly}
  \hat{d}_t &= \sum_{k=1}^{\beta_d} w_t^{(k)} \hat{d}_t^{(k)} \nonumber\\
            &= \sum_{k=1}^{\beta_d} \sum_{p \in \M_k} w_{t,p} \, \hat{d}_{t,p_k},
\end{align}
where $\M_k$ is the set of all leaf nodes in model $k$, and $p_k \in \Pre(p')$ is the longest prefix to the string $p'$ in the $k$th model, i.e., $p_k \triangleq \Pre(p') \cap \M_k$. Let $\Pre(p')=\{\nu'_1,\dots,\nu'_{d+1}\}$ denote the set of all prefixes to string $p'$. We then observe that the regressors of the nodes $\nu'_j \in \Pre(p')$ will be sufficient to obtain the final estimate of our algorithm. Therefore, we only consider the estimates of $O(d)$ nodes.

In order to further simplify the final estimate in \eqref{eq:final_estimate_doubly}, we first let $\Sp_d(p) \triangleq \{\acute{p} \in \N_d \, | \, \Pre(\acute{p}) = p\}$, i.e., $\Sp_d(p)$ denotes the set of all nodes of a depth-$d$ tree, whose set of prefixes include the node $p$. As an example, for a depth-$2$ tree, we have $\Sp(0) = \{0, 00, 01\}$. We then define a function $\rho(p,\acute{p})$ for arbitrary two nodes $p, \acute{p} \in \N_d$, as the number of models having both $p$ and $\acute{p}$ as its leaf nodes. Trivially, if $\acute{p} = p$, then $\rho(p,p) = \gamma_d(l(p))$. If $p \neq \acute{p}$, then letting $\bar{p}$ denote the longest prefix to both $p$ and $\acute{p}$, i.e., the longest string in $\Pre(p) \cap \Pre(\acute{p})$, we obtain
\begin{equation}\label{eq:rho_definition}
 \rho(p,\acute{p}) \hspace{-0.1cm} \triangleq \hspace{-0.1cm}
 \begin{cases}
   \gamma_d(l(p)) ,& \hspace{-0.3cm} \text{if } p = \acute{p} \\
   \frac{\gamma_d(l(p)) \gamma_{d-l(\bar{p})-1}(l(\acute{p})-l(\bar{p})-1)}{\beta_{d-l(\bar{p})-1}} ,& \hspace{-0.3cm} \text{if } p_1 \notin \Pre(\acute{p}) \cup \Sp_d(\acute{p}) \\
   0 ,& \hspace{-0.3cm} \text{otherwise}
 \end{cases} \hspace{-0.1cm}.
\end{equation}
Since $l(\bar{p})+1 \leq l(p), l(\acute{p})$ from the definition of the tree, we naturally have $\rho(p,\acute{p}) = \rho(\acute{p},p)$.

Now turning our attention back to \eqref{eq:final_estimate_doubly} and considering the definition in \eqref{eq:rho_definition}, we notice that the number of occurrences of the product $w_{t,p} \, \hat{d}_{t,p_k}$ in $\hat{d}_t$ is given by $\rho(p,p_k)$. Hence, the combination weight of the estimate of the node $p$ at time $t$ can be calculated as follows
\begin{equation}\label{eq:kappa}
  \kappa_{t,p} \triangleq \sum_{\acute{p} \in \N_d} \rho(p,\acute{p}) w_{t,\acute{p}}.
\end{equation}
Then, the final estimate of our algorithm becomes
\begin{equation}\label{eq:Final_Estimate_Hard}
  \hat{d}_t = \sum_{\nu'_j \in \Pre(p')} \kappa_{t,\nu'_j} \, \hat{d}_{t,\nu'_j}.
\end{equation}
We emphasize that the estimate of our algorithm given in \eqref{eq:Final_Estimate_Hard} achieves the exact same result with $\hat{d}_t = \vw_t^T \vhd_t$ with a computational complexity of $O(d2^d)$. Hence, the proof is concluded. \hfill $\square$

\section{Regressor Space Partitioning via Adaptive Soft Separator Functions}\label{sec:soft_partitioning}
In this section, the sequential regressors (as described in Section
\ref{sec:specific_partition}) for all partitions in the doubly
exponential tree class are combined when soft separation functions are
used, i.e., $s_t=\left(1+e^{\vx_t^T\vt_t}\right)^{-1}$, where $\vx_t \in
\Real^{m+1}$ is the extended regressor vector and $\vt_t$ is the
extended direction vector. By using soft separator functions, we train
the corresponding region boundaries, i.e., the structure of the tree.

As in Section \ref{sec:hard_partitioning}, for $\beta_d$ different
models that are embedded within a depth-$d$ tree, we introduce the
algorithm (given in Algorithm \ref{alg:Theorem2}) achieving asymptotically the
same cumulative squared regression error as the optimal combination of
the best adaptive models. The algorithm is constructed in the proof of
the Theorem 3.

The computational complexity of the algorithm of Theorem 3 is
$O(m4^d)$ whereas it achieves the performance of the best combination
of $O(2^{2^d})$ different ``adaptive'' regressors that partitions the
$m$-dimensional regressor space. The computational complexity of the
first algorithm was $O(d2^d)$, however, it was unable to learn the
region boundaries of the regressor space. In this case since we are
using soft separator functions, we need to consider the
cross-correlation of every node estimate and node weight, whereas in
the previous case there we were only considering the cross-correlation of
the estimates of the prefixes of the node $p \in \Leaf_d$ such that
$\vx_t \in \R_p$ and the weights of every node. This change transforms
the computational complexity from $O(d2^d)$ to $O(4^d)$. Moreover, for
all inner nodes a soft separator function is defined. In order to
update the region boundaries of the partitions, we have to update the
direction vector $\vt_t$ of size $m$ since $\vx_t \in
\Real^m$. Therefore, considering the cross-correlation of the final
estimates of every node, we get a computational complexity of
$O(m4^d)$.

{\bf Theorem 3: }{\em Let $\sdt$ and $\svxt$ be arbitrary, bounded, and real-valued sequences. The algorithm $\hat{d}_t$ given in Algorithm \ref{alg:Theorem2} when applied these sequences yields
\begin{equation}\label{eq:Theorem2}
   \sum_{t=1}^n \big{(} d_t - \hat{d}_t \big{)}^2 - \min_{\vw \in \Real^{\beta_d}} \sum_{t=1}^n \big{(} d_t - \vw^T\vhd_t \big{)}^2 \leq O \big{(} \ln(n) \big{)},
\end{equation}
for all $n$, when $e_t^2(\vw)$ is strongly convex $\forall t$, where $\vhd_t=[\hat{d}_t^{(1)},\dots,\hat{d}_t^{(\beta_d)}]^T$ and $\hat{d}_t^{(k)}$ represents the estimate of $d_t$ at time $t$ for the adaptive model $k=1,\dots,\beta_d$.}

This theorem implies that our algorithm (given in Algorithm
\ref{alg:Theorem2}), asymptotically achieves the performance of the
best linear combination of the $O(2^{2^d})$ different adaptive models
that can be represented using a depth-$d$ tree with a computational
complexity $O(m4^d)$. We emphasize that while constructing the
algorithm, we refrain from any statistical assumptions on the
underlying data, and our algorithm works for any sequence of $\sdt$
with an arbitrary length of $n$. Furthermore, one can use this
algorithm to learn the region boundaries and then feed this information to the first algorithm to reduce computational complexity.

\subsection{Outline of the Proof of Theorem 3 and Construction of Algorithm 2}
The proof of the upper bound in Theorem 3 follows similar lines to the
proof of upper bound in Theorem 1, therefore is omitted. In this proof, we provide the detailed algorithmic description and highlight the computational complexity differences.

According to the same labeling operation we presented in Section \ref{sec:hard_partitioning}, the final estimate of the $k$th model at time $t$ can be found as follows
\[
\hat{d}^{(k)} = \sum_{p \in \M_k} \hd_{t,p}.
\]
Similarly, the weight of the $k$th model is given by
\[
w_t^{(k)} = \sum_{p \in \M_k} w_{t,p}.
\]

Since we use soft separator functions, we have $\hd_{t,p} > 0$ and without introducing any approximations, the final estimate of our algorithm is given as follows
\[
\hat{d}_t = \sum_{k=1}^{\beta_d} \left\{ \left( \sum_{p \in \M_k} w_{t,p} \right)  \left( \sum_{p \in \M_k} \hd_{t,p} \right) \right\}.
\]
Here, we observe that for arbitrary two nodes $p, \acute{p} \in \N_d$, the product $w_{t,p} \hd_{t,\acute{p}}$ appears $\rho(p,\acute{p})$ times in $\hat{d}_t$, where $\rho(p,\acute{p})$ is the number of models having both $p$ and $\acute{p}$ as its leaf nodes (as we previously defined in \eqref{eq:rho_definition}). Hence, according to the notation derived in \eqref{eq:rho_definition} and \eqref{eq:kappa}, we obtain the final estimate of our algorithm as follows
\begin{equation}\label{eq:Final_Estimate_Soft}
  \hat{d}_t = \sum_{p \in \N_d} \kappa_{t,p} \, \hd_{t,p}.
\end{equation}
Note that \eqref{eq:Final_Estimate_Soft} is equal to $\hat{d}_t = \vw_t^T \vhd_t$ with a computational complexity of $O(4^d)$.

%\begin{table}
\begin{algorithm}
    \caption{Decision Adaptive Tree (DAT) Regressor}\label{alg:Theorem2}
    \begin{algorithmic}[1]
        \FOR{$t=1$ \TO $n$}
            \STATE $\hat{d}_t \Leftarrow 0$
            \FORALL{$p \in \N_d - \Leaf_d$}
                \STATE $s_{t,p} \Leftarrow s^+ + (1-2s^+)/(1+e^{\vx_t^T \vt_{t,p}})$
            \ENDFOR
            \FORALL{$p \in \Leaf_d$}
                \STATE $\hat{d}_{t,p} \Leftarrow \vv_{t,p}^T \vx_t$
                \STATE $\alpha_{t,p} \Leftarrow 1$
                \FOR{$i=1$ \TO $l(p)$}
                    \STATE $\alpha_{t,p} \Leftarrow \alpha_{t,p} s_{t,\nu_i}^{q_i}$
                \ENDFOR
                \STATE $\hd_{t,p} \Leftarrow \alpha_{t,p} \hat{d}_{t,p}$
                \STATE $\kappa_{t,p} \Leftarrow \gamma_d\big{(}l(p)\big{)} w_{t,p}$
                \FORALL{$\acute{p} \in \N_d-(\Pre(p) \cup \Sp_d(p))$}
                    \STATE $\bar{p} \Leftarrow \tilde{p} \in \Pre(p) \cap \Pre(\acute{p}) : l(\tilde{p}) = \left| \Pre(p) \cap \Pre(\acute{p}) \right| -1$
                    \STATE $\kappa_{t,p} \Leftarrow \kappa_{t,p} + \frac{ \gamma_d\big{(}l(p)\big{)} \gamma_{d-l(\bar{p})-1}\big{(}l(\acute{p})-l(\bar{p})-1\big{)} }{ \beta_{d-l(\bar{p})-1} } w_{t,\acute{p}}$
                \ENDFOR
            \STATE $\hat{d}_t \Leftarrow \hat{d}_t + \kappa_{t,p} \hd_{t,p}$
            \ENDFOR
            \STATE $e_t \Leftarrow d_t - \hat{d}_t$
            \FORALL{$p \in \Leaf_d$}
                \STATE $\vv_{t+1,p} \Leftarrow \vv_{t,p} + \mu_t e_t \alpha_{t,p} \vx_t$
                \STATE $w_{t+1,p} \Leftarrow w_{t,p} + \mu_t e_t \hd_{t,p}$
            \ENDFOR
            \FORALL{$p \in \N_d - \Leaf_d$}
                \STATE $\sigma_{t,p} \Leftarrow 0$
                \FORALL{$\acute{p} \in \Sp_d(p0)$}
                    \STATE $\sigma_{t,p} \Leftarrow \sigma_{t,p} + \kappa_{t,\acute{p}} \frac{\hd_{t,\acute{p}}}{s_{t,p}}$
                \ENDFOR
                \FORALL{$\acute{p} \in \Sp_d(p1)$}
                    \STATE $\sigma_{t,p} \Leftarrow \sigma_{t,p} - \kappa_{t,\acute{p}} \frac{\hd_{t,\acute{p}}}{1-s_{t,p}}$
                \ENDFOR
                \STATE $\vt_{t+1,p} \Leftarrow \vt_{t,p} - \eta_t e_t \sigma_{t,p} s_{t,p} (1-s_{t,p}) \vx_t$
            \ENDFOR
        \ENDFOR
    \end{algorithmic}
\end{algorithm}
%  \caption{Algorithm II: Online Piecewise Linear Regressor for Adaptive and Soft Partitioning of the Regressor Space}\label{alg:Theorem2}
%\end{table}

Unlike Section \ref{sec:hard_partitioning}, in which each model has a fixed partitioning of the regressor space, here, we define the regressor models with adaptive partitions. For this, we use a stochastic gradient descent update
\begin{equation}\label{eq:inner_node_update}
  \vt_{t+1,p} = \vt_{t,p} - \frac{1}{2} \eta_t \nabla e_t^2(\vt_{t,p}),
\end{equation}
for all nodes $p \in \N_d - \Leaf_d$, where $\eta_t$ is the learning rate of the region boundaries and $\nabla e_t^2(\vt_{t,p})$ is the derivative of $e_t^2(\vt_{t,p})$ with respect to $\vt_{t,p}$. After some algebra, we obtain
\begin{align}\label{eq:theta_update}
  \vt_{t+1,p} &= \vt_{t,p} + \eta_t e_t \, \frac{\partial \hat{d}_t}{\partial s_{t,p}} \, \frac{\partial s_{t,p}}{\partial \vt_{t,p}}, \nonumber\\
    &= \vt_{t,p} + \eta_t e_t \left\{ \sum_{\acute{p} \in \N_d} \kappa_{t,\acute{p}} \frac{\partial \hd_{t,\acute{p}}}{\partial s_{t,p}} \right\} \frac{\partial s_{t,p}}{\partial \vt_{t,p}} \nonumber\\
    &= \vt_{t,p} + \eta_t e_t \left\{ \sum_{q=0}^1 \sum_{\acute{p} \in \Sp_d(pq)} (-1)^q \kappa_{t,\acute{p}} \frac{\hd_{t,\acute{p}}}{s_{t,p}^q} \right\} \frac{\partial s_{t,p}}{\partial \vt_{t,p}},
\end{align}
where we use the logistic regression classifier as our separator function, i.e., $s_{t,p} = \left( 1 + \exp(\vx_t^T \vt_{t,p}) \right)^{-1}$. Therefore, we have
\begin{align}\label{eq:separator_derivative}
  \frac{\partial s_{t,p}}{\partial \vt_{t,p}} &= - \left( 1 + \exp(\vx_t^T \vt_{t,p}) \right)^{-2} \exp(\vx_t^T \vt_{t,p}) \vx_t \nonumber\\
    &= - s_{t,p} (1-s_{t,p}) \vx_t.
\end{align}
Note that other separator functions can also be used in a similar way by simply calculating the gradient with respect to the extended direction vector and plugging in \eqref{eq:theta_update} and  \eqref{eq:separator_derivative}.

We emphasize that $\nabla e_t^2(\vt_{t,p})$ includes the product of $s_{t,p}$ and $1-s_{t,p}$ terms, hence in order not to slow down the learning rate of our algorithm, we may restrict $s^+ \leq  |s_t| \leq 1-s^+$ for some $0 < s^+ < 0.5$. According to this restriction, we define the separator functions as follows
\[
s_t = s^+ + \frac{1-2s^+}{1+e^{\vx_t^T\vt_t}}.
\]

According to the update rule in \eqref{eq:theta_update}, the computational complexity of the introduced algorithm results in $O(m4^d)$. This concludes the outline of the proof and the construction of the algorithm. \hfill $\square$

\subsection{Selection of the Learning Rates}
We emphasize that the learning rate $\mu_t$ can be set according to the similar studies in the literature \cite{sayed_book, Hazan} or considering the application requirements. However, for the introduced algorithm to work smoothly, we expect the region boundaries to converge faster than the node weights, therefore, we conventionally choose the learning rate to update the region boundaries as $\eta_t = \mu_t/(s^+(1-s^+))$. Experimentally, we observed that different choices of $\eta_t$ also yields acceptable performance, however, we note that when updating $\vt_{t,p}$, we have the multiplication term $s_{t,p}(1-s_{t,p})$, which significantly decreases the steps taken at each time $t$. Therefore, in order to compensate for it, such a selection is reasonable.

On the other hand, for stability purposes, one can consider to put an upper bound on the steps at each time $t$. When $\vx_t$ is sufficiently away from the region boundaries $s_{t,p}$, it is either close to $s^+$ or $1-s^+$. However, when $\vx_t$ falls right on a region boundary, we have $s_{t,p}=0.5$, which results in an approximately $25$ times greater step than the expected one, when $s^+=0.01$. This issue is further exacerbated when $\vx_t$ falls on the boundary of multiple region crossings, e.g., say $\vx_t = [0, \, 0]^T$ when we have the four quadrants as the four regions (leaf nodes) of the depth-$2$ tree. In such a scenario, one can observe a $25^d$ times greater step than expected, which may significantly perturb the stability of the algorithm. That is why, two alternate solutions can be proposed: {\em 1)} a reasonable threshold (e.g., $10s^+(1-s^+)$)) over the steps can be embedded when $s^+$ is small (or equivalently, a regularization constant can be embedded), {\em 2)} $s^+$ can be sufficiently increased according to the depth of the tree. Throughout the experiments, we used the first approach.

\subsection{Selection of the Depth of the Tree}
In many real life applications, we do not know how the true data is generated, therefore, the accurate selection of the depth of the decision tree is usually a difficult problem. For instance, if the desired data is generated from a piecewise linear model, then in order for the conventional approaches that use a fixed tree structure (i.e., fixed partitioning of the regressor space) to perfectly estimate the data, they need to perfectly guess the underlying partitions in hindsight. Otherwise, in order to capture the salient characteristics of the desired data, the depth of the tree should be increased to infinity. Hence, the performance of such algorithms significantly varies according to the initial partitioning of the regressor space, which makes it harder to decide how to select the depth of the tree.

On the other hand, the introduced algorithm adapts its region boundaries to minimize the final regression error. Therefore, even if the initial partitioning of the regressor space is not accurate, our algorithm will learn to the locally optimal partitioning of the regressor space for any given depth $d$. In this sense, one can select the depth of the decision tree by only considering the computational complexity issues of the application.

\section{Simulations}\label{sec:Simulations}
In this section, we illustrate the performance of our algorithms under different scenarios with respect to various methods. We first consider the regression of a signal generated by a piecewise linear model when the underlying partition of the model corresponds to one of the partitions represented by the tree. We then consider the case when the partitioning does not match any partition represented by the tree to demonstrate the region-learning performance of the introduced algorithm. We also illustrate the performance of our algorithms in underfitting and overfitting (in terms of the depth of the tree) scenarios. We then consider the prediction of two benchmark chaotic processes: the Lorenz attractor and the Henon map. Finally, we illustrate the merits of our algorithm using benchmark data sets (both real and synthetic) such as California housing \cite{delve,keel,ltorgo}, elevators \cite{delve}, kinematics \cite{keel}, pumadyn \cite{keel}, and bank \cite{ltorgo} (which will be explained in detail in Subsection \ref{ssec:real}).

Throughout this section, ``DFT'' represents the decision fixed tree regressor (i.e., Algorithm 1) and ``DAT'' represents the decision adaptive tree regressor (i.e., Algorithm 2). Similarly, ``CTW'' represents the context tree weighting algorithm of \cite{CTW}, ``OBR'' represents the optimal batch regressor, ``VF'' represents the truncated Volterra filter \cite{volterra}, ``LF'' represents the simple linear filter, ``B-SAF'' and ``CR-SAF'' represent the Beizer and the Catmul-Rom spline adaptive filter of \cite{saf}, respectively, ``FNF'' and ``EMFNF'' represent the Fourier and even mirror Fourier nonlinear filter of \cite{fnf}, respectively. Finally, ``GKR'' represents the Gaussian-Kernel regressor and it is constructed using $p$ node regressors, say $\hat{d}_{t,1}, \dots, \hat{d}_{t,p}$, and a fixed Gaussian mixture weighting (that is selected according to the underlying sequence in hindsight), giving
\[
\hat{d}_t = \sum_{i=1}^p f\left(\vx_t;\vmu_i,\vsigma_i\right) \hat{d}_{t,i},
\]
where $\hat{d}_{t,i} = \vv_{t,i}^T \vx_t$ and
\[
f\left(\vx_t;\vmu_i,\vsigma_i\right) \triangleq \frac{1}{2\pi\sqrt{\left|\vsigma_i\right|}} e^{-\frac{1}{2}(\vx_t-\vmu_i)^T \vsigma_i^{-1} (\vx_t-\vmu_i)},
\]
for all $i = 1,\dots,p$.

For a fair performance comparison, in the corresponding experiments in Subsections \ref{ssec:chaotic} and \ref{ssec:real}, the desired data and the regressor vectors are normalized between $[-1,1]$ since the satisfactory performance of the several algorithms require the knowledge on the upper bounds (such as the B-SAF and the CR-SAF) and some require these upper bounds to be between $[-1,1]$ (such as the FNF and the EMFNF). Moreover, in the corresponding experiments in Subsections \ref{ssec:matched}, \ref{ssec:mismatched}, and \ref{ssec:fitting}, the desired data and the regressor vectors are normalized between $[-1,1]$ for the VF, the FNF, and the EMFNF due to the aforementioned reason. The regression errors of these algorithms are then scaled back to their original values for a fair comparison.

Considering the illustrated examples in the respective papers \cite{saf,fnf,CTW}, the orders of the FNF and the EMFNF are set to $3$ for the experiments in Subsections \ref{ssec:matched}, \ref{ssec:mismatched}, and \ref{ssec:fitting}, $2$ for the experiments in Subsection \ref{ssec:chaotic}, and $1$ for the experiments in Subsection \ref{ssec:real}. The order of the VF is set to $2$ for all experiments, except for the California housing experiment, in which it is set to $3$. Similarly, the depth of the tree of the DAT algorithm is set to $2$ for all experiments, except for the California housing experiment, in which it is set to $3$. The depths of the trees of the DFT and the CTW algorithms are set to $2$ for all experiments. For the tree based algorithms, the regressor space is initially partitioned by the direction vectors $\vt_{t,p} = [\theta_{t,p}^{(1)},\dots,\theta_{t,p}^{(m)}]^T$ for all nodes $p \in \N_d - \Leaf_d$, where $\theta_{t,p}^{(i)} = -1$ if $i \equiv l(p) \pmod{d}$, e.g., when $d=m=2$, we have the four quadrants as the four leaf nodes of the tree. Finally, we used cubic B-SAF and CR-SAF algorithms, whose number of knots are set to $21$ for all experiments. We emphasize that both these parameters and the learning rates of these algorithms are selected to give equal rate of performance and convergence.

\subsection{Computational Complexities}
\begin{table}
  \centering
  \resizebox{.75\columnwidth}{!}{
  \begin{tabular}{| c | c |}
    \hline
    Algorithm & Computational Complexity \\ \hline
    DFT       & $O\left( md2^d \right)$ \\ \hline
    DAT       & $O\left( m4^d \right)$ \\ \hline
    CTW       & $O\left( md \right)$ \\ \hline
    GKR       & $O\left( m2^d \right)$ \\ \hline
    VF        & $O\left( m^r \right)$ \\ \hline
    B-SAF     & $O\left( mr^2 \right)$ \\ \hline
    CR-SAF    & $O\left( mr^2\right)$ \\ \hline
    FNF       & $O\left( (mr)^r \right)$ \\ \hline
    EMFNF     & $O\left( m^r \right)$ \\ \hline
  \end{tabular}
  }
  \caption{Comparison of the computational complexities of the proposed algorithms. In the table, $m$ represents the dimensionality of the regressor space, $d$ represents the depth of the trees in the respective algorithms, and $r$ represents the order of the corresponding filters and algorithms.}\label{tab:complexity}
\end{table}

As can be observed from Table \ref{tab:complexity}, among the tree based algorithms that partition the regressor space, the CTW algorithm has the smallest complexity since at each time $t$, it only associates the regressor vector $\vx_t$ with $O(d)$ nodes (the leaf node $\vx_t$ has fallen into and all its prefixes) and their individual weights. The DFT algorithm also considers the same $O(d)$ nodes on the tree, but in addition, it calculates the weight of the each node with respect to the rest of the nodes, i.e., it correlates $O(d)$ nodes with all the $O(2^d)$ nodes. The DAT algorithm, however, estimates the data with respect to the correlation of all the nodes, one another, which results in a computational complexity of $O(4^d)$. In order for the Gaussian-Kernel Regressor (GKR) to achieve a comparable nonlinear modeling power, it should have $2^d$ mass points, which results in a computational complexity of $O(m2^d)$.

On the other hand, the filters such as the VF, the FNF, and the EMFNF introduce the nonlinearity by directly considering the $r$th (and up to $r$th) powers of the entries of the regressor vector. In many practical applications, such methods cannot be applied due to the high dimensionality of the regressor space. Therefore, the algorithms such as the B-SAF and the CR-SAF are introduced to decrease the high computational complexity of such approaches. However, as can be observed from our simulation results, the introduced algorithm significantly outperforms its competitors in various benchmark problems.

The algorithms such as the VF, the FNF, and the EMFNF have more than enough number of basis functions, which result in a significantly slower and parameter dependent convergence performance with respect to the other algorithms. On the other hand, the performances of the algorithms such as the B-SAF, the CR-SAF, and the CTW algorithm are highly dependent on the underlying setting that generates the desired signal. Furthermore, for all these algorithms to yield satisfactory results, prior knowledge on the desired signals and the regressor vectors is needed. The introduced algorithms, on the other hand, do not rely on any prior knowledge, and still outperform their competitors.

\subsection{Matched Partitions}\label{ssec:matched}
\begin{figure}
  \centering
  \includegraphics[width=0.5\textwidth]{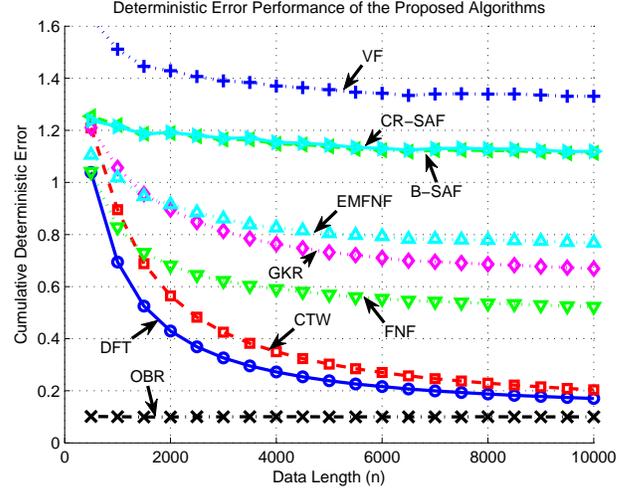}\\
  \caption{Regression error performances for the second order piecewise linear model in \eqref{eq:simulation_1} averaged over 10 trials.}\label{fig:error_matched}
\end{figure}

\begin{figure*}
    \centering
    \begin{subfigure}[b]{0.49\textwidth}
        \centering
        \includegraphics[width=\textwidth]{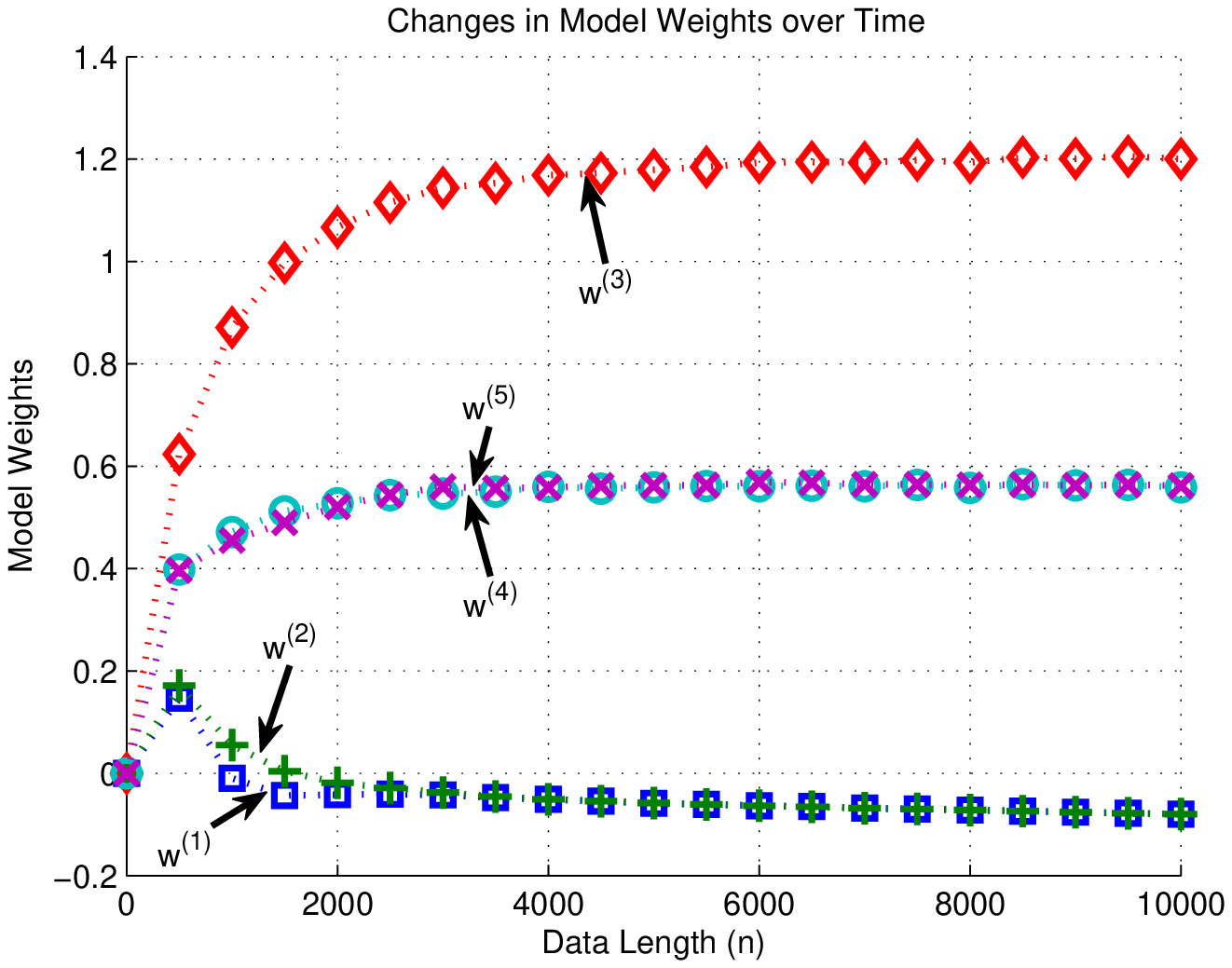}\\
        \caption{}\label{fig:model_weights_matched}
    \end{subfigure}
    \begin{subfigure}[b]{0.49\textwidth}
        \centering
        \includegraphics[width=\textwidth]{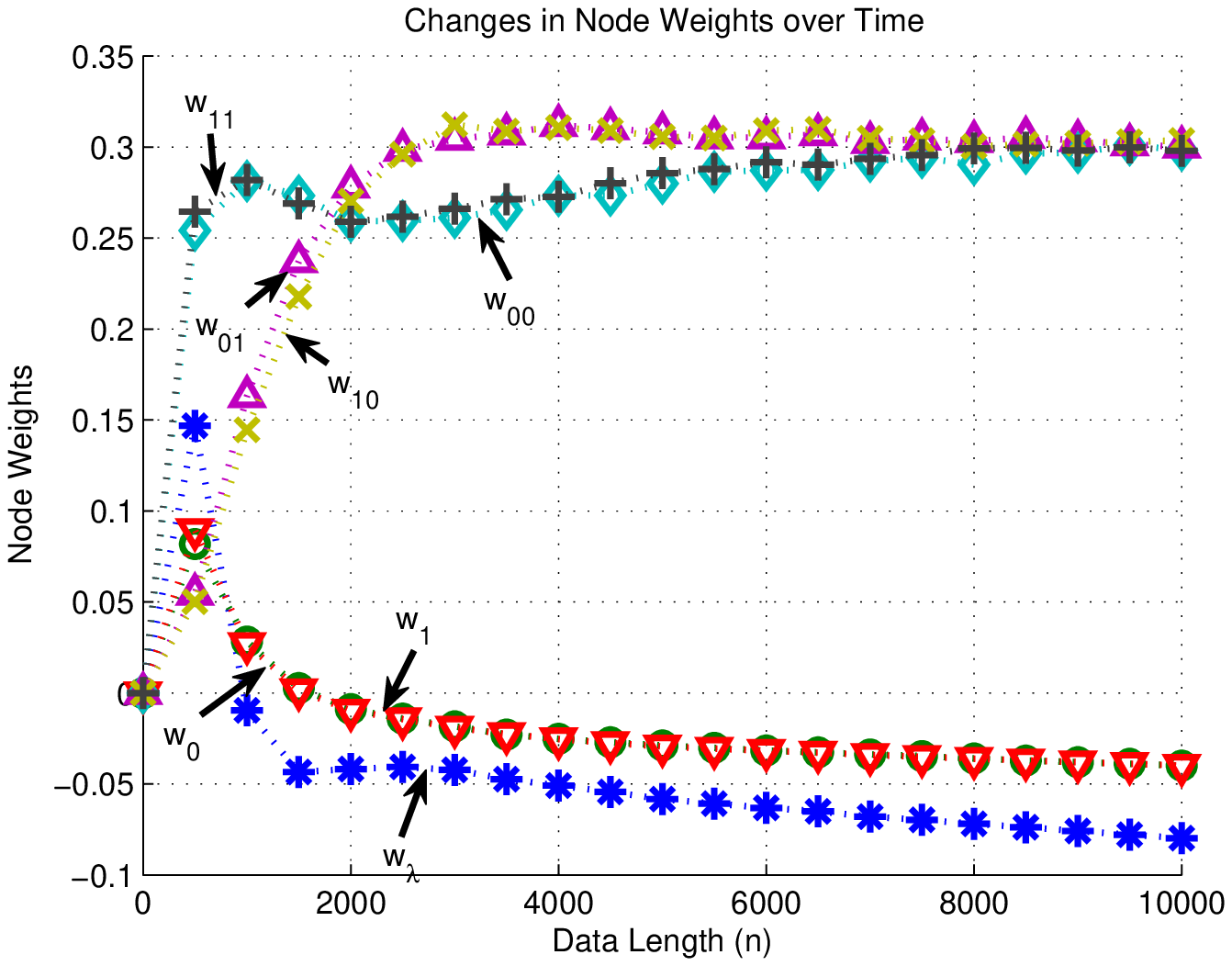}\\
        \caption{}\label{fig:node_weights_matched}
    \end{subfigure}
    \caption{Progress of (a) the model weights and (b) the node weights averaged over 10 trials for the DFT algorithm. Note that the model weights do not sum up to 1.}\label{fig:model&node_weights_matched}
\end{figure*}

In this subsection, we consider the case where the desired data is generated by a piecewise linear model that matches with the initial partitioning of the tree based algorithms. Specifically, the desired signal is generated by the following piecewise linear model
\begin{equation}\label{eq:simulation_1}
  d_t =
    \begin{cases}
        \vw^T \vx_t + \pi_t ,& \text{if } \vp_0^T \vx_t \geq 0 \text{ and } \vp_1^T \vx_t \geq 0 \\
        -\vw^T \vx_t + \pi_t ,& \text{if } \vp_0^T \vx_t \geq 0 \text{ and } \vp_1^T \vx_t < 0 \\
        -\vw^T \vx_t + \pi_t ,& \text{if } \vp_0^T \vx_t < 0 \text{ and } \vp_1^T \vx_t \geq 0 \\
        \vw^T \vx_t + \pi_t ,& \text{if } \vp_0^T \vx_t < 0 \text{ and } \vp_1^T \vx_t < 0 \\
    \end{cases},
\end{equation}
where $\vw = [1, \, 1]^T$, $\vp_0 = [1, \, 0]^T$, $\vp_1 = [0, \, 1]^T$, $\vx_t = [x_{1,t}, \, x_{2,t}]^T$, $\pi_t$ is a sample function from a zero mean white Gaussian process with variance $0.1$, $x_{1,t}$ and $x_{2,t}$ are sample functions of a jointly Gaussian process of mean $[0, \, 0]^T$ and variance $\vI_2$. The desired data at time $t$ is denoted as $d_t$ whereas the extended regressor vector is $\vx_t = [x_{1,t}, x_{2,t}, 1]^T$, i.e., $x_1$ represents the first dimension and $x_2$ the second dimension.

For this scenario, the learning rates are set to $0.005$ for the DFT algorithm, the FNF, and the CTW algorithm, $0.025$ for the B-SAF and the CR-SAF, $0.05$ for the VF and the EMFNF, $1$ for the GKR. Moreover, for the GKR, $\vmu_i = 1.2 \times [(-1)^{\lfloor (i-1)/2 \rfloor}, \, (-1)^{i}]^T$ and $\vsigma_i = 1.2 \times \vec{I}_2$ for $i=1,\dots,4$, are set to exactly match the underlying partitioning that generates the desired data.

In Fig. \ref{fig:error_matched}, we demonstrate the time accumulated regression error of the proposed algorithms averaged over $10$ trials. Since the desired data is generated by a highly nonlinear piecewise model, the algorithms such as the GKR, the FNF, the EMFNF, the B-SAF, and the CR-SAF cannot capture the salient characteristics of the data. These algorithms yield satisfactory results only if the desired data is generated by a smooth nonlinear function of the regressor vector. In this scenario, however, we have high nonlinearity and discontinuity, which makes the algorithms such as the DFT and the CTW appealing.

Comparing the DFT and the CTW algorithms, we can observe that even though the partitioning of the tree perfectly matches with the underlying partition in \eqref{eq:simulation_1}, the learning performance of the DFT algorithm significantly outperforms the CTW algorithm especially for short data records. As commented in the text, this is expected since the context-tree weighting method enforces the sum of the model weights to be $1$, however, the introduced algorithms have no such restrictions. As seen in Fig. \ref{fig:model_weights_matched}, the model weights sum up to $2.1604$ instead of $1$. Moreover, in the CTW algorithm all model weights are ``forced'' to be nonnegative whereas in our algorithm model weights can also be negative as seen in Fig. \ref{fig:model_weights_matched}. In Fig. \ref{fig:node_weights_matched}, the individual node weights are presented. We observe that the nodes (i.e., regions) that directly match with the underlying partition that generates the desired data have higher weights whereas the weights of the other nodes decrease. We also point out that although the tree based algorithms \cite{Takimoto1,Takimoto2,CTW} need a priori information, such as an upper bound on the desired data, for a successful operation, whereas the introduced algorithm has no such requirements.

\subsection{Mismatched Partitions}\label{ssec:mismatched}
In this subsection, we consider the case where the desired data is generated by a piecewise linear model that mismatches with the initial partitioning of the tree based algorithms. Specifically, the desired signal is generated by the following piecewise linear model
\begin{equation}\label{eq:simulation_2}
  d_t =
    \begin{cases}
        \vw^T \vx_t + \pi_t ,& \text{if } \vp_0^T \vx_t \geq 0.5 \text{ and } \vp_1^T \vx_t \geq 1 \\
        -\vw^T \vx_t + \pi_t ,& \text{if } \vp_0^T \vx_t \geq 0.5 \text{ and } \vp_1^T \vx_t < 1 \\
        -\vw^T \vx_t + \pi_t ,& \text{if } \vp_0^T \vx_t < 0.5 \text{ and } \vp_2^T \vx_t \geq -1 \\
        \vw^T \vx_t + \pi_t ,& \text{if } \vp_0^T \vx_t < 0.5 \text{ and } \vp_2^T \vx_t < -1 \\
    \end{cases},
\end{equation}
where $\vw = [1, \, 1]^T$, $\vp_0 = [4, \, -1]^T$, $\vp_1 = [1, \, 1]^T$, $\vp_2 = [1, \, 2]^T$, $\vx_t = [x_{1,t}, \, x_{2,t}]^T$, $\pi_t$ is a sample function from a zero mean white Gaussian process with variance $0.1$, $x_{1,t}$ and $x_{2,t}$ are sample functions of a jointly Gaussian process of mean $[0, \, 0]^T$ and variance $\vI_2$. The learning rates are set to $0.005$ for the DFT, the DAT, and the CTW algorithms, $0.1$ for the FNF, $0.025$ for the B-SAF and the CR-SAF, $0.05$ for the EMFNF and the VF. Moreover, in order to match the underlying partition, the mass points of the GKR are set to $\vmu_1 = [1.4565 ,\, 1.0203]^T$, $\vmu_2 = [0.6203 ,\, -0.4565]^T$, $\vmu_3 = [-0.5013 ,\, 0.5903]^T$, and $\vmu_4 = [-1.0903 ,\, -1.0013]^T$ with the same covariance matrix in the previous example.

\begin{figure}
  \centering
  \includegraphics[width=.5\textwidth]{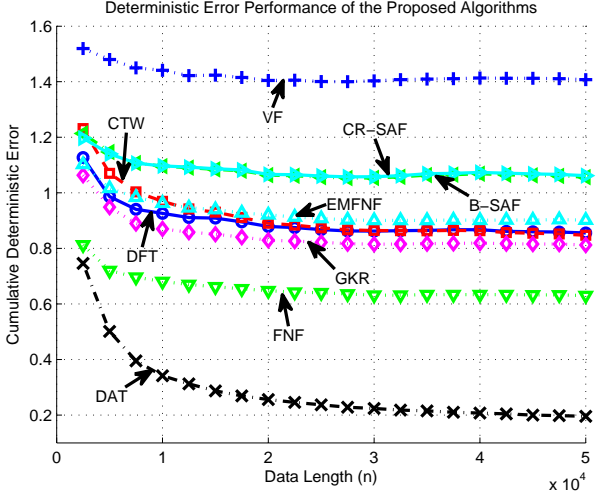}\\
  \caption{Regression error performances for the second order piecewise linear model in \eqref{eq:simulation_2}.}\label{fig:error_mismatched}
\end{figure}

Fig. \ref{fig:error_mismatched} shows the normalized time accumulated regression error of the proposed algorithms. We emphasize that the DAT algorithm achieves a better error performance compared to its competitors. Comparing Fig. \ref{fig:error_matched} and Fig. \ref{fig:error_mismatched}, one can observe the degradation in the performances of the DFT and the CTW algorithms. This shows the importance of the initial partitioning of the regressor space for tree based algorithms to yield a satisfactory performance. Comparing the same figures, one can also observe that the rest of the algorithms performs almost similar to the previous scenario.

\begin{figure*}
    \centering
    \begin{subfigure}[b]{0.3\textwidth}
        \centering
        \includegraphics[scale=0.4]{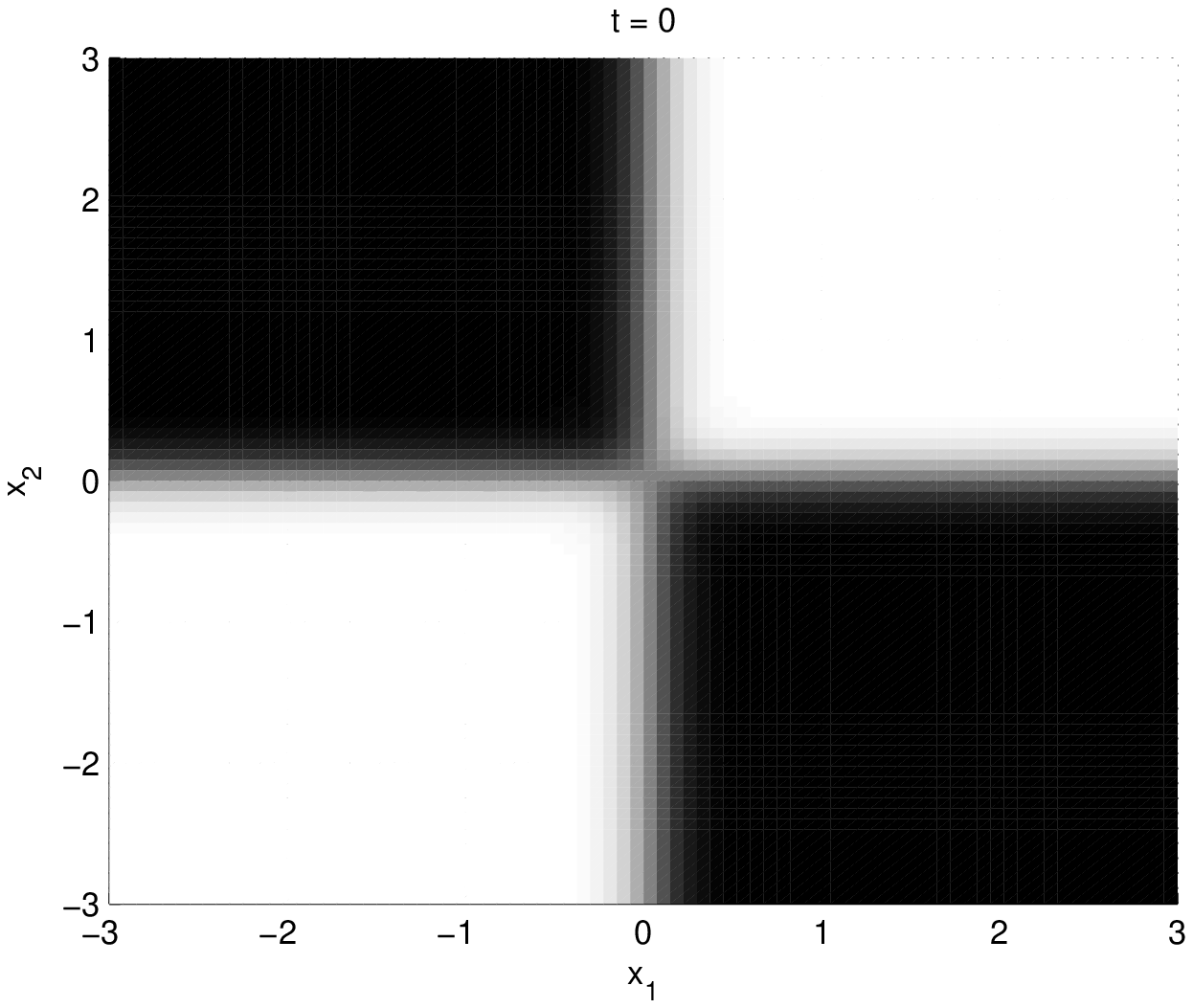}
        \label{fig:regions_1}
    \end{subfigure}
    \begin{subfigure}[b]{0.3\textwidth}
        \centering
        \includegraphics[scale=0.4]{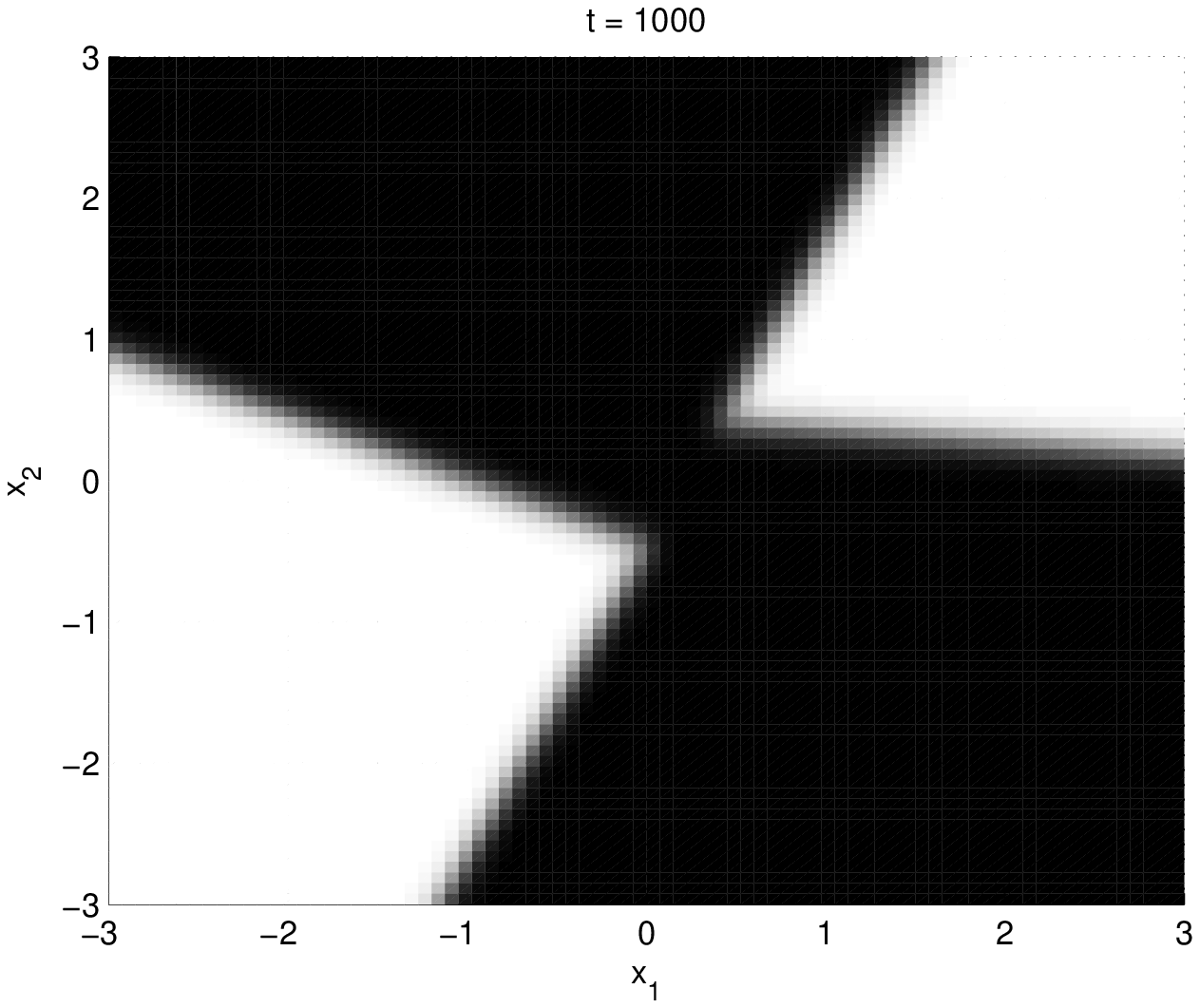}
        \label{fig:regions_2}
    \end{subfigure}
    \begin{subfigure}[b]{0.3\textwidth}
        \centering
        \includegraphics[scale=0.4]{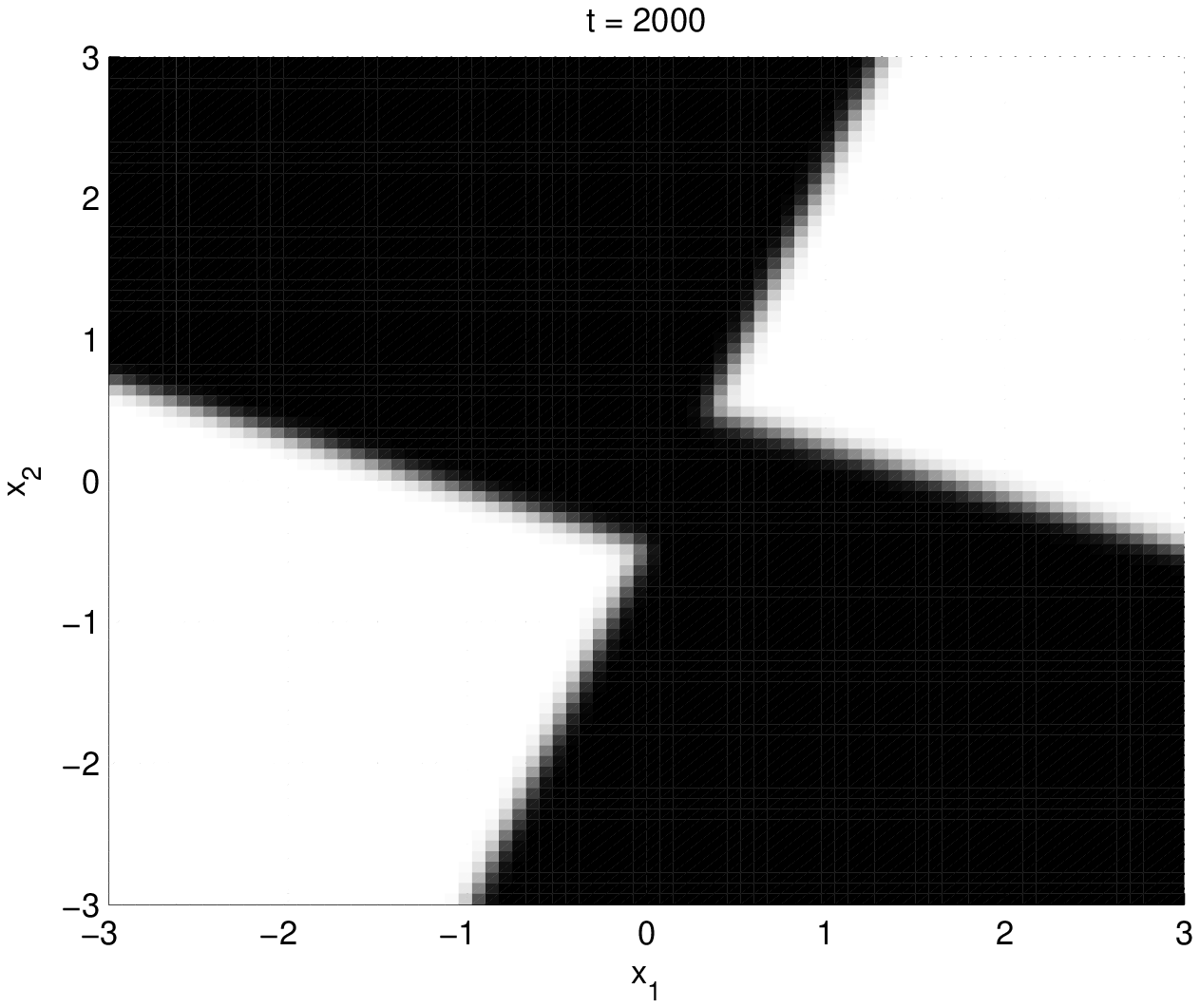}
        \label{fig:regions_3}
    \end{subfigure}
    \linebreak
    \linebreak
    \begin{subfigure}[b]{0.3\textwidth}
        \centering
        \includegraphics[scale=0.4]{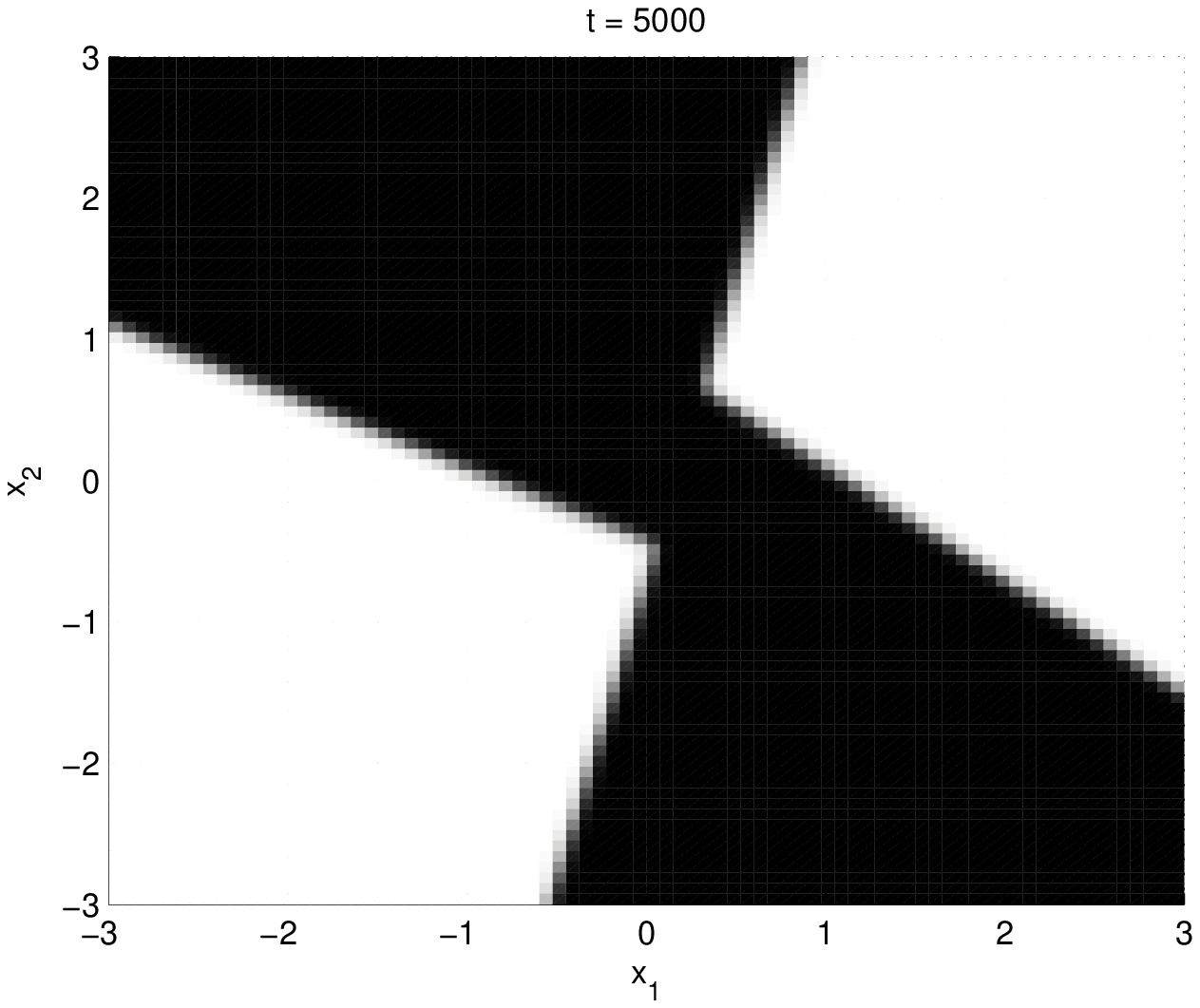}
        \label{fig:regions_4}
    \end{subfigure}
    \begin{subfigure}[b]{0.3\textwidth}
        \centering
        \includegraphics[scale=0.4]{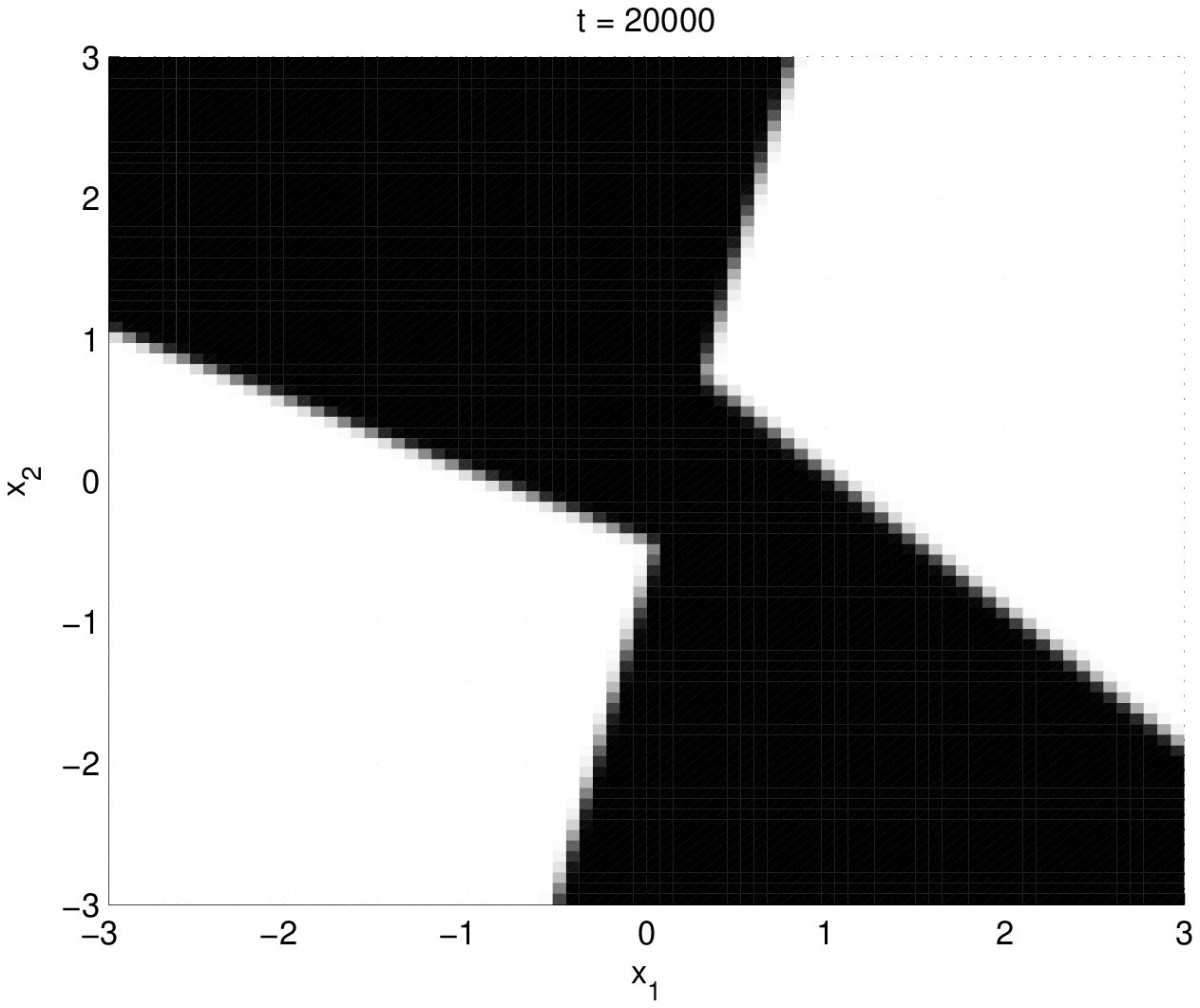}
        \label{fig:regions_5}
    \end{subfigure}
    \begin{subfigure}[b]{0.3\textwidth}
        \centering
        \includegraphics[scale=0.4]{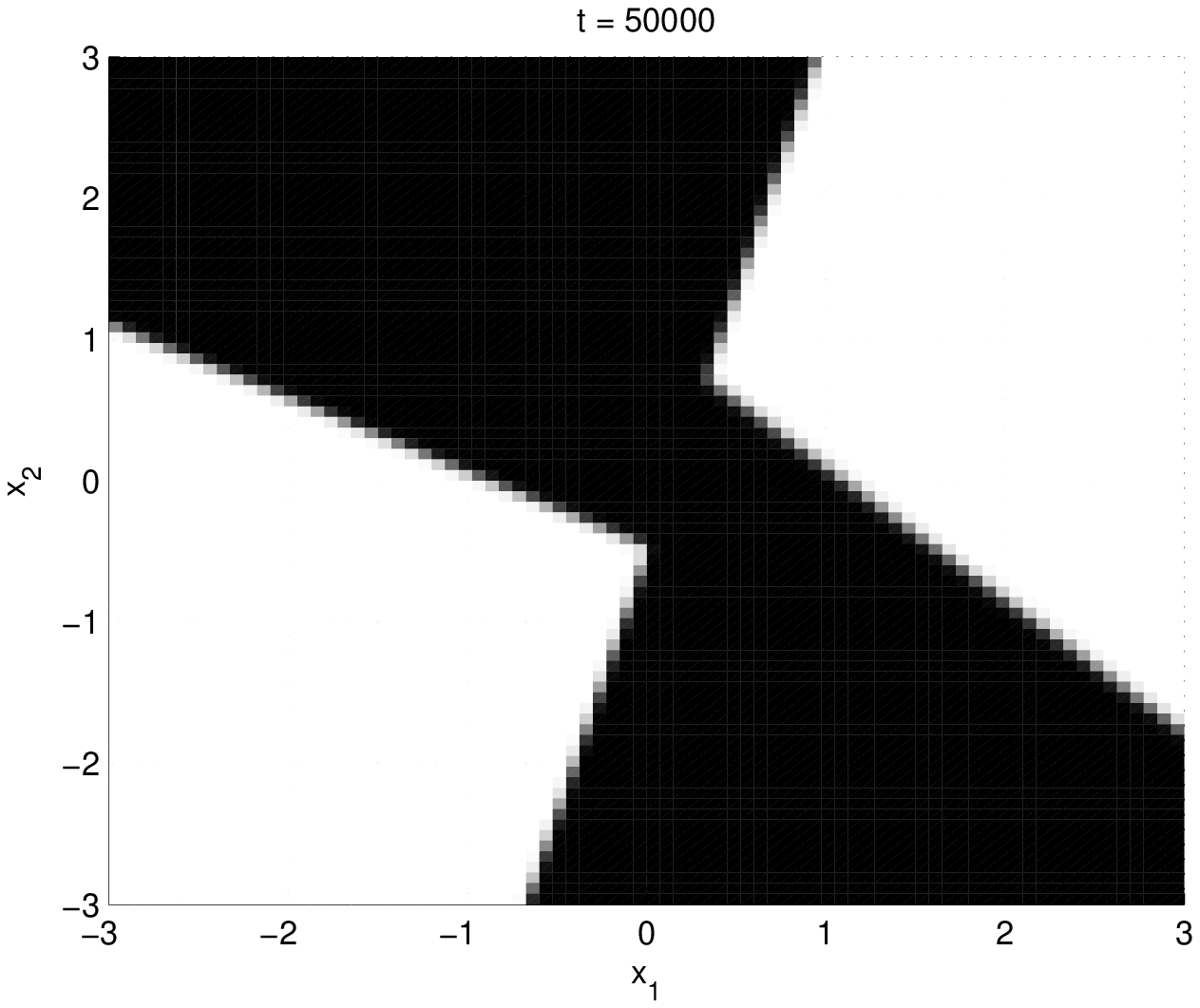}
        \label{fig:regions_6}
    \end{subfigure}
    \caption{Changes in the boundaries of the leaf nodes of the depth-$2$ tree of the DAT algorithm for $t = 0, 1000, 2000, 5000, 20000, 50000$. The separator functions adaptively learn the boundaries of the piecewise linear model in \eqref{eq:simulation_2}.}\label{fig:regions}
\end{figure*}

The DFT and the CTW algorithms converge to the best batch regressor having the predetermined leaf nodes (i.e., the best regressor having the four quadrants of two dimensional space as its leaf nodes). However that regressor is sub-optimal since the underlying data is generated using another constellation, hence their time accumulated regression error is always lower bounded by $O(1)$ compared to the global optimal regressor. The DAT algorithm, on the other hand, adapts its region boundaries and captures the underlying unevenly rotated and shifted regressor space partitioning, perfectly. Fig. \ref{fig:regions} shows how our algorithm updates its separator functions and illustrates the nonlinear modeling power of the introduced DAT algorithm.

\begin{figure*}
    \centering
    \begin{subfigure}[b]{0.49\textwidth}
        \centering
        \includegraphics[width=1\textwidth]{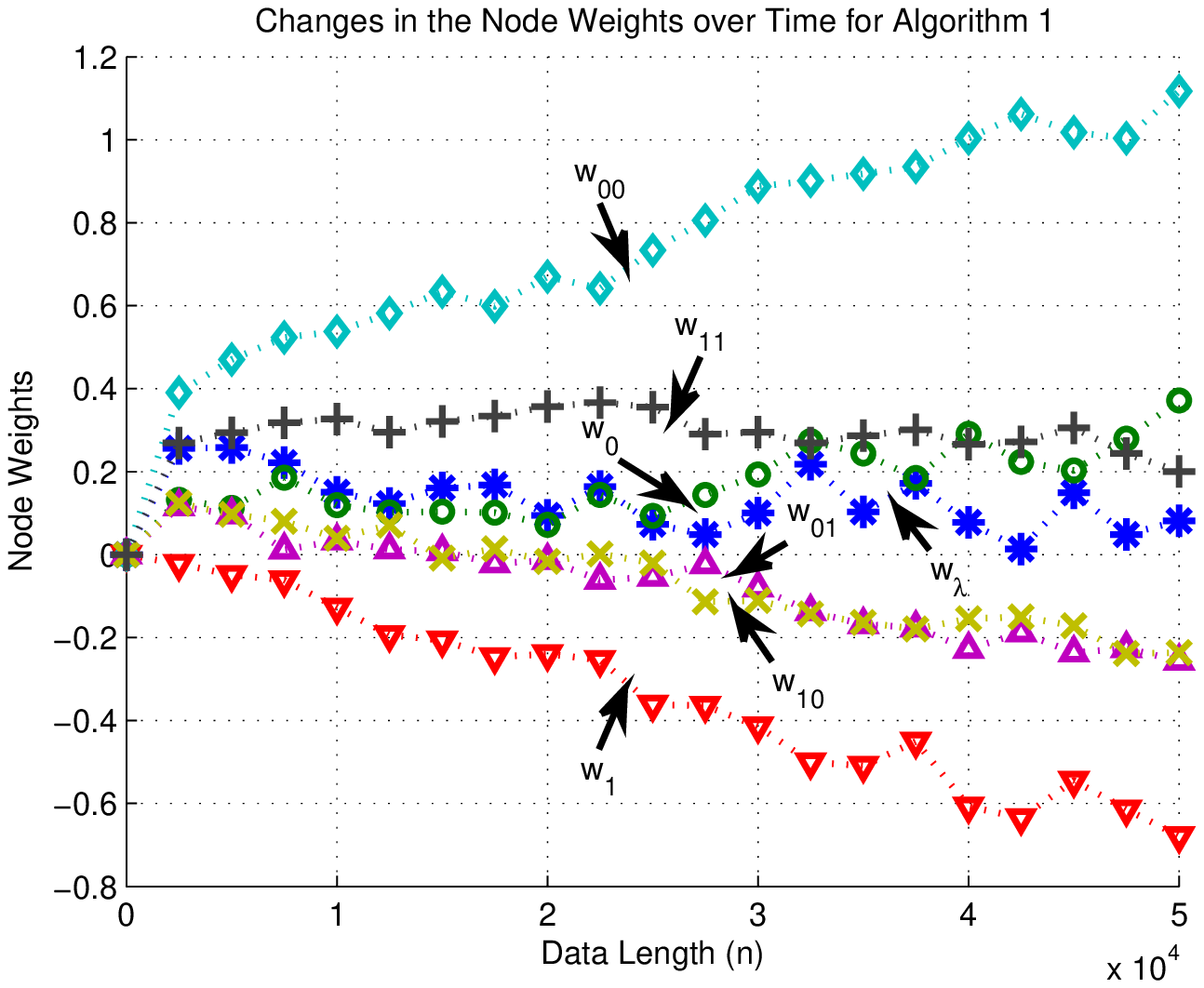}\\
        \caption{}\label{fig:node_weights_hard_2}
    \end{subfigure}
    \begin{subfigure}[b]{0.49\textwidth}
        \centering
        \includegraphics[width=1\textwidth]{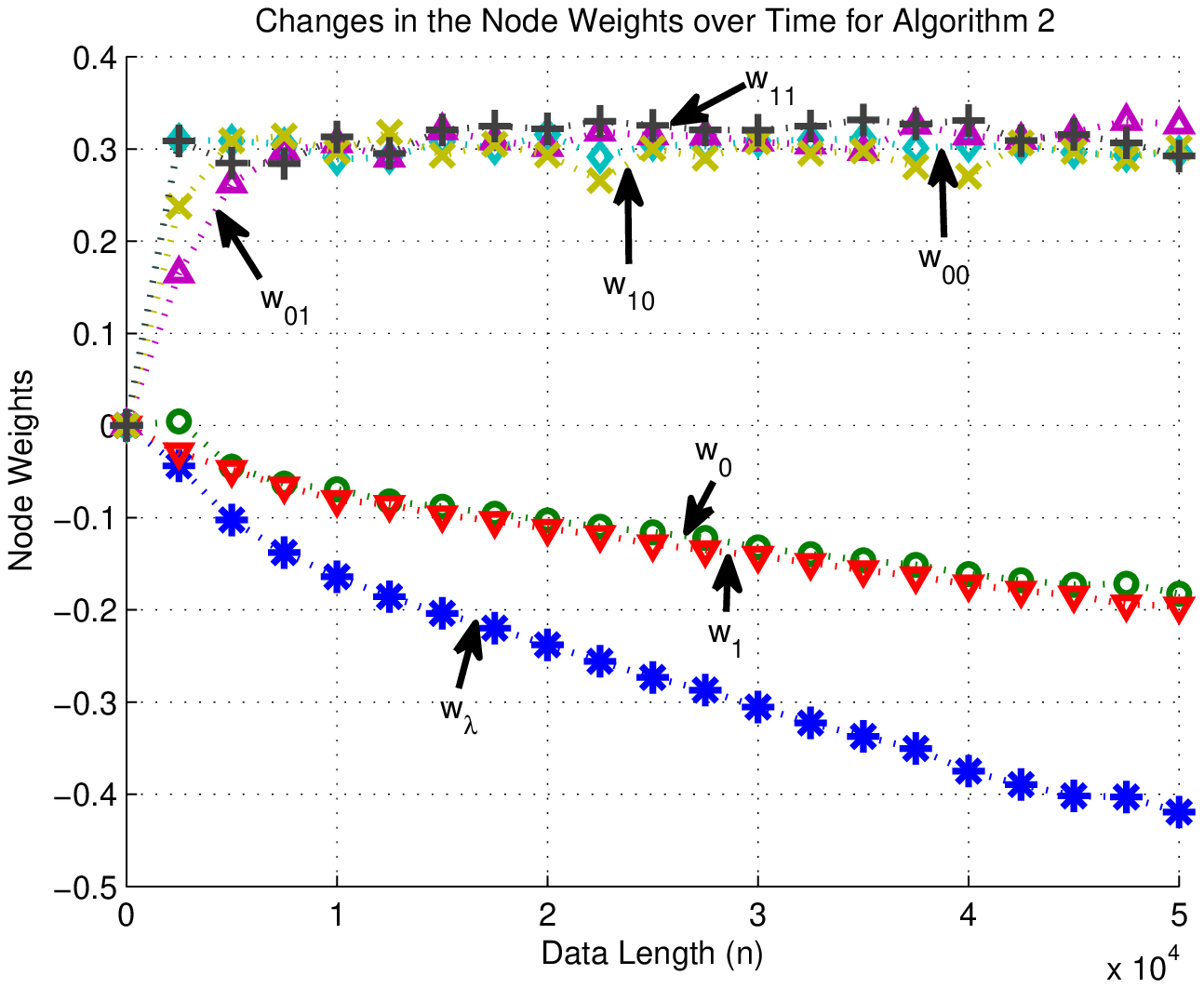}\\
        \caption{}\label{fig:node_weights_soft_2}
    \end{subfigure}
    \caption{Progress of the node weights for the piecewise linear model in \eqref{eq:simulation_2} for (a) the DFT algorithm and (b) the DAT algorithm.}\label{fig:node_weights_2}
\end{figure*}

We also present the node weights for the DFT and the DAT algorithms in Fig. \ref{fig:node_weights_hard_2} and Fig. \ref{fig:node_weights_soft_2}, respectively. In Fig. \ref{fig:node_weights_hard_2}, we can observe that the DFT algorithm cannot estimate the underlying data accurately, hence its node weights show unstable behavior. On the other hand, as can be observed from Fig. \ref{fig:node_weights_soft_2}, the DAT algorithm learns the optimal node weights as the region boundaries are learned. In this manner, the DAT algorithm achieves a significantly superior performance with respect to its competitors.

\subsection{Mismatched Partitions with Overfitting \& Underfitting}\label{ssec:fitting}
In this subsection, we consider two cases (and perform two experiments), where the desired data is generated by a piecewise linear model that mismatches with the initial partitioning of the tree based algorithms, where the depth of the tree overfits or underfits the underlying piecewise model. In the first set of experiments, we consider that the data is generated from a first order piecewise linear model, for which using a depth-$1$ tree is sufficient to capture the salient characteristics of the data. In the second set of experiments, we consider that the data is generated from a third order piecewise linear model, for which it is necessary to use a depth-$3$ tree to perfectly estimate the data.

The first order piecewise linear model is defined as
\begin{equation}\label{eq:simulation_overfitting}
  d_t =
    \begin{cases}
        \vw^T \vx_t + \pi_t ,& \text{if } \vp_0^T \vx_t \geq 0.5 \\
        -\vw^T \vx_t + \pi_t ,& \text{if } \vp_0^T \vx_t < 0.5  \\
    \end{cases},
\end{equation}
and the third order piecewise linear model is defined as
\begin{equation}\label{eq:simulation_underfitting}
  d_t \hspace{-.1cm} = \hspace{-.15cm}
    \begin{cases}
        \hspace{-.1cm} \vw^T \vx_t \hspace{-.1cm} + \hspace{-.1cm} \pi_t ,& \hspace{-.25cm} \text{if } \vp_0^T \vx_t \geq 0.5 \text{, } \vp_1^T \vx_t \geq 1 \text{, } \vp_3^T \vx_t \geq 0.5  \\
        \hspace{-.1cm} -\vw^T \vx_t \hspace{-.1cm} + \hspace{-.1cm} \pi_t ,& \hspace{-.25cm} \text{if } \vp_0^T \vx_t \geq 0.5 \text{, } \vp_1^T \vx_t \geq 1 \text{, } \vp_3^T \vx_t < 0.5  \\
        \hspace{-.1cm} \vw^T \vx_t \hspace{-.1cm} + \hspace{-.1cm} \pi_t ,& \hspace{-.25cm} \text{if } \vp_0^T \vx_t \geq 0.5 \text{, } \vp_1^T \vx_t < 1 \text{, } \vp_4^T \vx_t \geq 0.5  \\
        \hspace{-.1cm} -\vw^T \vx_t \hspace{-.1cm} + \hspace{-.1cm} \pi_t ,& \hspace{-.25cm} \text{if } \vp_0^T \vx_t \geq 0.5 \text{, } \vp_1^T \vx_t < 1 \text{, } \vp_4^T \vx_t < 0.5  \\
        \hspace{-.1cm} \vw^T \vx_t \hspace{-.1cm} + \hspace{-.1cm} \pi_t ,& \hspace{-.25cm} \text{if } \vp_0^T \vx_t < 0.5 \text{, } \vp_2^T \vx_t < 0.5 \text{, } \vp_5^T \vx_t < 0.5  \\
        \hspace{-.1cm} -\vw^T \vx_t \hspace{-.1cm} + \hspace{-.1cm} \pi_t ,& \hspace{-.25cm} \text{if } \vp_0^T \vx_t < 0.5 \text{, } \vp_2^T \vx_t < 0.5 \text{, } \vp_5^T \vx_t \geq 0.5  \\
        \hspace{-.1cm} \vw^T \vx_t \hspace{-.1cm} + \hspace{-.1cm} \pi_t ,& \hspace{-.25cm} \text{if } \vp_0^T \vx_t < 0.5 \text{, } \vp_2^T \vx_t \geq 0.5 \text{, } \vp_6^T \vx_t < 0.5  \\
        \hspace{-.1cm} -\vw^T \vx_t \hspace{-.1cm} + \hspace{-.1cm} \pi_t ,& \hspace{-.25cm} \text{if } \vp_0^T \vx_t < 0.5 \text{, } \vp_2^T \vx_t \geq 0.5 \text{, } \vp_6^T \vx_t \geq 0.5  \\
    \end{cases},
\end{equation}
where $\vw = [1, \, 1]^T$, $\vp_0 = [4, \, -1]^T$, $\vp_1 = [1, \, 1]^T$, $\vp_2 = [-1, \, -2]^T$, $\vp_3 = [0, \, 1]^T$, $\vp_4 = [1, \, 0]^T$, $\vp_5 = [-1, \, 0]^T$, $\vp_6 = [0, \, -1]^T$, $\vx_t = [x_{1,t}, \, x_{2,t}]^T$, $\pi_t$ is a sample function from a zero mean white Gaussian process with variance $0.1$, $x_{1,t}$ and $x_{2,t}$ are sample functions of a jointly Gaussian process of mean $[0, \, 0]^T$ and variance $\vI_2$. The learning rates are set to $0.005$ for the DFT, the DAT, and the CTW algorithms, $0.05$ for the EMFNF, $0.01$ for the B-SAF, the CR-SAF, and the FNF, $0.5$ for the VF, and $1$ for the GKR, where the parameters of the GKR are set to the same values in the previous example.

\begin{figure*}
    \centering
    \begin{subfigure}[b]{0.49\textwidth}
        \centering
        \includegraphics[width=\textwidth]{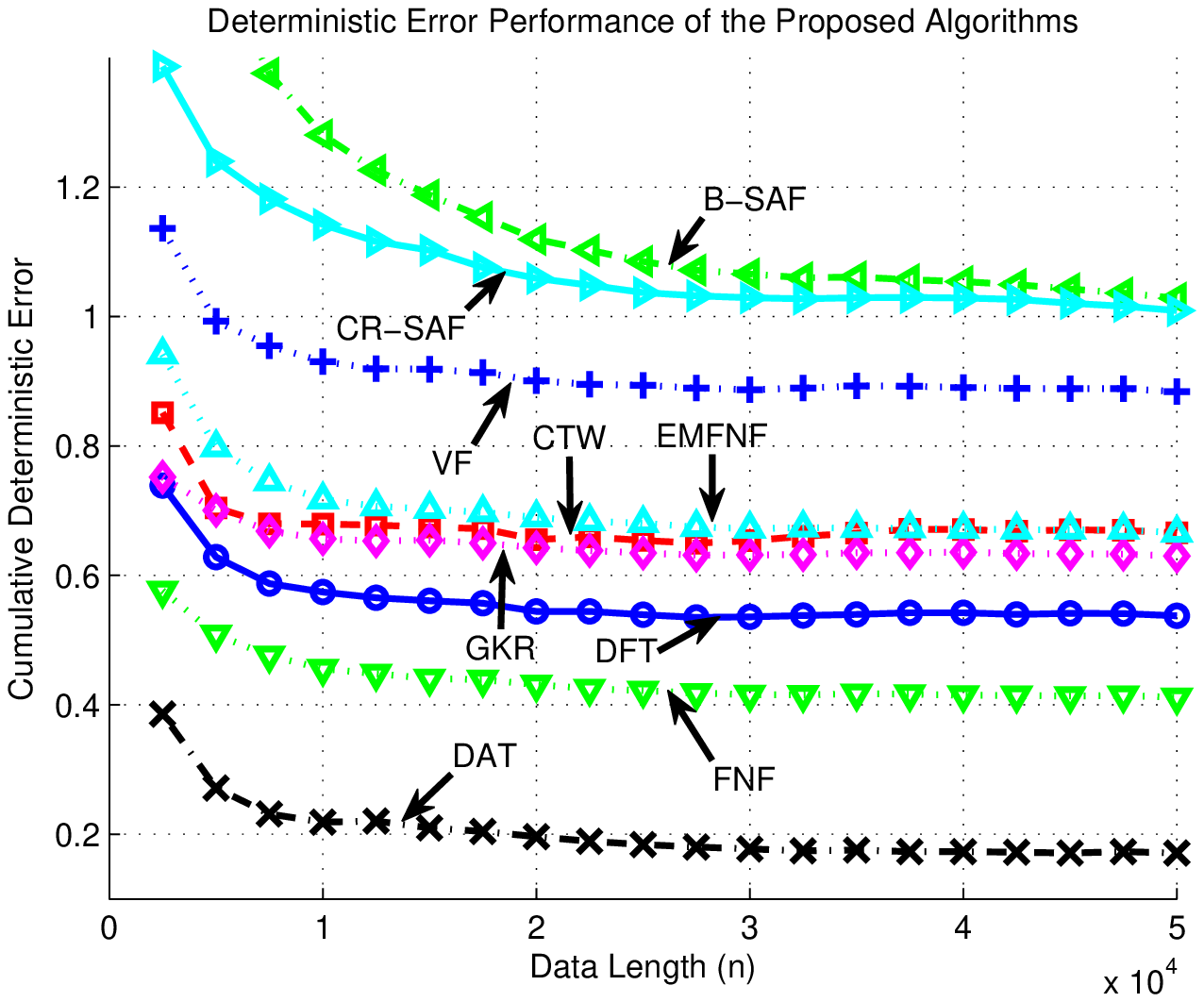}\\
        \caption{}\label{fig:error_overfitting}
    \end{subfigure}
    \begin{subfigure}[b]{0.49\textwidth}
        \centering
        \includegraphics[width=\textwidth]{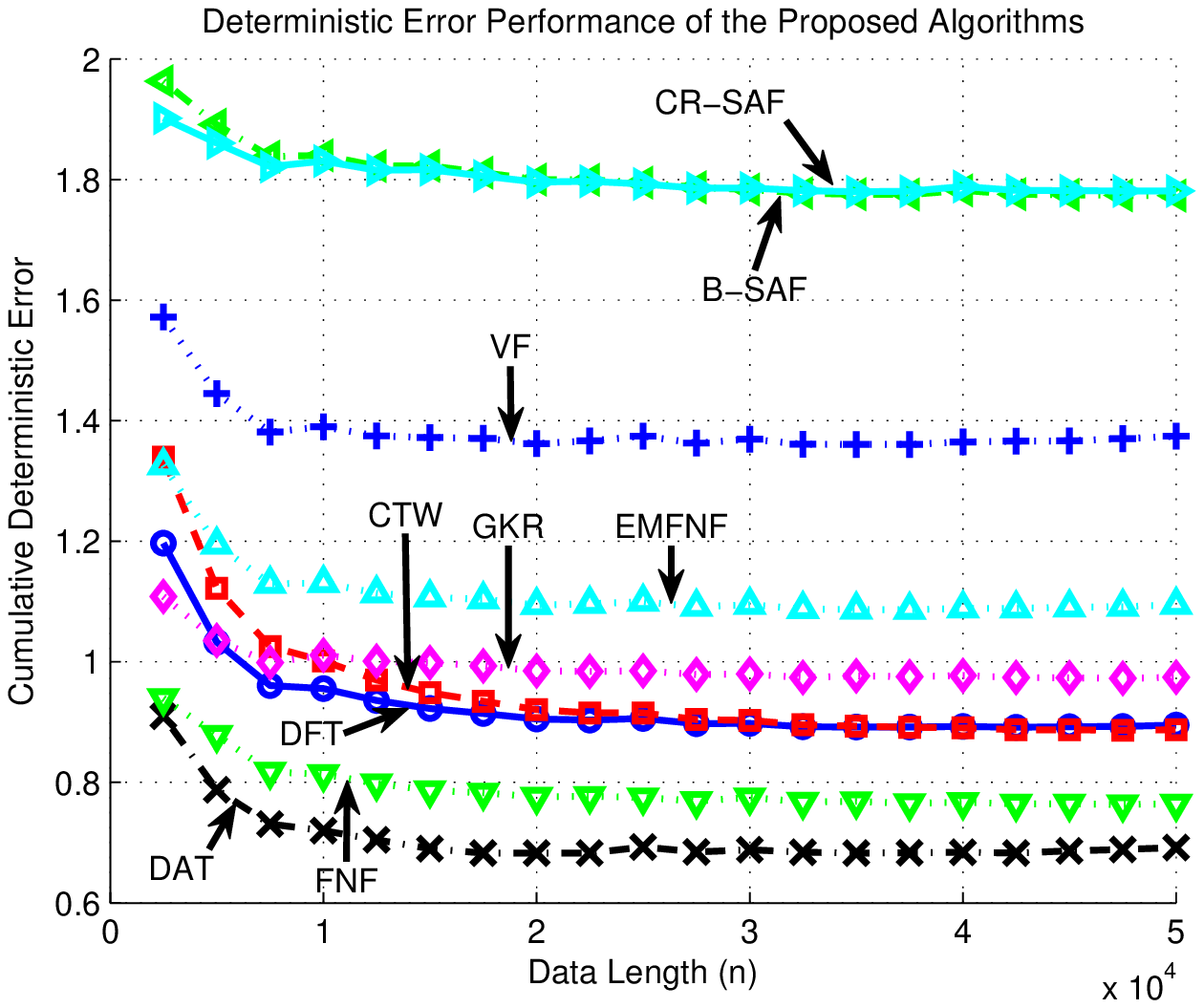}\\
        \caption{}\label{fig:error_underfitting}
    \end{subfigure}
    \caption{Regression error performances for (a) the first order piecewise linear model in \eqref{eq:simulation_overfitting} (b) the third order piecewise linear model in \eqref{eq:simulation_underfitting}.}\label{fig:error_fitting}
\end{figure*}

We present the normalized regression errors of the proposed algorithms in Fig. \ref{fig:error_fitting}. Fig. \ref{fig:error_overfitting} shows the performances of the algorithms in the overfitting scenario, where the desired data is generated by the first order piecewise linear model in \eqref{eq:simulation_overfitting}. Similarly, Fig. \ref{fig:error_underfitting} shows the performances of the algorithms in the underfitting scenario, where the desired data is generated by the third order piecewise linear model in \eqref{eq:simulation_underfitting}. From the figures, it is observed that the DAT algorithm outperforms its competitors by learning the optimal partitioning for the given depth, which illustrates the power of the introduced algorithm under possible mismatches in terms of $d$.

\subsection{Chaotic Signals}\label{ssec:chaotic}
In this subsection, we illustrate the performance of our algorithm when estimating a chaotic data generated by {\em i)} the Henon map and {\em ii)} the Lorenz attractor \cite{lorenz}.

First, we consider a zero-mean sequence generated by the Henon map, a chaotic process given by
\begin{equation}\label{eq:simulation_3}
  d_t = 1 - \zeta \, d_{t-1}^2 + \eta \, d_{t-2},
\end{equation}
and known to exhibit chaotic behavior for the values of $\zeta = 1.4$ and $\eta = 0.3$. The desired data at time $t$ is denoted as $d_t$ whereas the extended regressor vector is $\vx_t = [d_{t-1}, d_{t-2}, 1]^T$, i.e., we consider a prediction framework. The learning rates are set to $0.025$ for the B-SAF and the CR-SAF algorithms, whereas it is $0.05$ for the rest.

\begin{figure}
  \centering
  \includegraphics[width=0.5\textwidth]{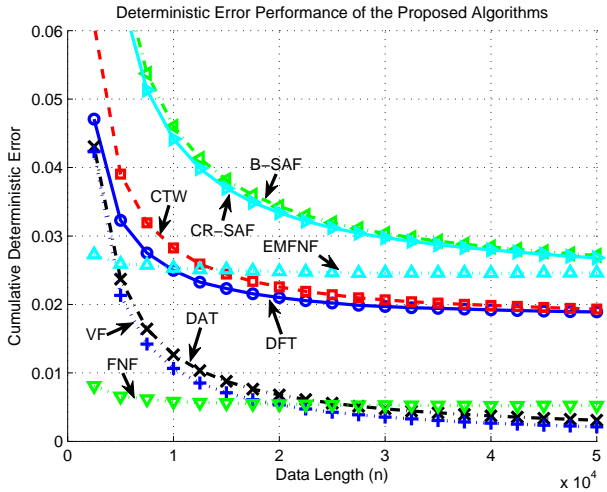}\\
  \caption{Regression error performances of the proposed algorithms for the chaotic process presented in \eqref{eq:simulation_3}.}\label{fig:error_henon}
\end{figure}

Fig. \ref{fig:error_henon} shows the normalized regression error performance of the proposed algorithms. One can observe that the algorithms whose basis functions do not include the necessary quadratic terms and the algorithms that rely on a fixed regressor space partitioning yield unsatisfactory performance. On the other hand, we emphasize that the VF can capture the salient characteristics of this chaotic process since its order is set to $2$. Similarly, the FNF can also learn the desired data since its basis functions can well approximate the chaotic process. The DAT algorithm, however, uses a piecewise linear modeling and still achieves the asymptotically same performance as the VF, while outperforming the FNF algorithm.

%\begin{figure}
%  \centering
%  \includegraphics[width=0.5\textwidth]{henon_map.eps}\\
%  \caption{The chaotic behavior of the Henon Map given in \eqref{eq:simulation_3}.}\label{fig:henon_map}
%\end{figure}

Second, we consider the chaotic signal set generated using the Lorenz attractor \cite{lorenz} that is defined by the following three discrete time equations:
\begin{align}
  x_t &= x_{t-1} + (\sigma(y-x))dt \label{eq:lorenz1}\\
  y_t &= y_{t-1} + (x_{t-1}(\rho-z_{t-1})-y_{t-1})dt \label{eq:lorenz2}\\
  z_t &= z_{t-1} + (x_{t-1}y_{t-1}-\beta z_{t-1})dt, \label{eq:lorenz3}
\end{align}
where we set $dt=0.01$, $\rho=28$, $\sigma=10$, and $\beta=8/3$ to generate the well-known chaotic solution of the Lorenz attractor. In the experiment, $x_t$ is selected as the desired data and the two dimensional region represented by $y_t, z_t$ is set as the regressor space, that is, we try to estimate $x_t$ with respect to $y_t$ and $z_t$. The learning rates are set to $0.01$ for all the algorithms.

\begin{figure}
  \centering
  \includegraphics[width=.5\textwidth]{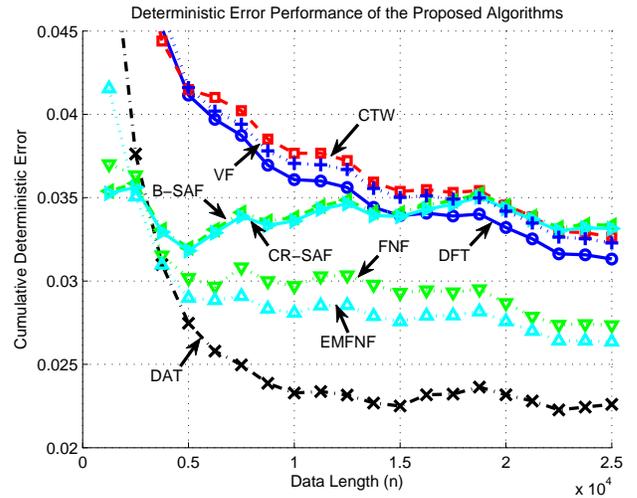}\\
  \caption{Regression error performances for the chaotic signal generated from the Lorenz attractor in \eqref{eq:lorenz1},\eqref{eq:lorenz2}, and \eqref{eq:lorenz3} with parameters $dt=0.01$, $\rho=28$, $\sigma=10$, and $\beta=8/3$.}\label{fig:error_lorenz}
\end{figure}

Fig. \ref{fig:error_lorenz} illustrates the nonlinear modeling power of the DAT algorithm even when estimating a highly nonlinear chaotic signal set. As can be observed from Fig. \ref{fig:error_lorenz}, the DAT algorithm significantly outperforms its competitors and achieves a superior error performance since it tunes its region boundaries to the optimal partitioning of the regressor space, whereas the performances of the other algorithms directly rely on the initial selection of the basis functions and/or tree structures and partitioning.

\subsection{Benchmark Real and Synthetic Data}\label{ssec:real}

\begin{table*}
  \centering
  \resizebox{1.8\columnwidth}{!}{
  \begin{tabular}{|l||*{7}{c|}} \hline
  \backslashbox[3em]{Data Sets \kern-2em}{\kern-2em Algorithms}
  &\makebox{DAT}&\makebox{LF}&\makebox{VF}&\makebox{FNF}&\makebox{EMFNF}&\makebox{B-SAF}&\makebox{CR-SAF} \\\hline\hline
  \hspace{.35cm} Kinematics & $0.0639$ & $0.0835$ & $0.0746$ & $0.0956$ & $0.0808$ & $0.1108$ & $0.1029$ \\\hline
  \hspace{.45cm} Elevators  & $0.0091$ & $0.0193$ & $0.0194$ & $0.0112$ & $0.0149$ & $0.0222$ & $0.0225$ \\\hline
  \hspace{.45cm} Pumadyn    & $0.0780$ & $0.0817$ & $0.0910$ & $0.0904$ & $0.0781$ & $0.0947$ & $0.0936$ \\\hline
  \hspace{.7cm} Bank      & $0.0511$ & $0.0739$ & $0.0804$ & $0.0544$ & $0.0533$ & $0.0764$ & $0.0891$ \\\hline
  \end{tabular}
  }
  \caption{Time accumulated normalized errors of the proposed algorithms. Each dimension of the data sets is normalized between $[-1,1]$.}\label{tab:error_table}
\end{table*}

\begin{figure}
  \centering
  \includegraphics[width=.5\textwidth]{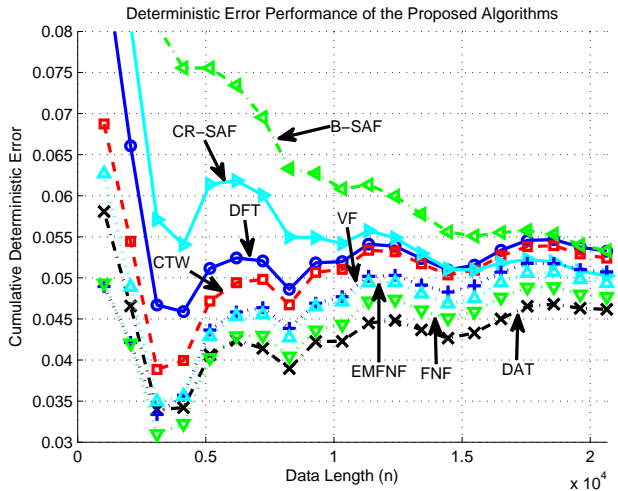}\\
  \caption{Regression error performances for the real data set: California housing - estimation of the median house prices in the California area using California housing database \cite{delve,keel,ltorgo}.}\label{fig:error_california_housing}
\end{figure}

In this subsection, we first consider the regression of a benchmark real-life problem that can be found in many data set repositories such as \cite{delve,keel,ltorgo}: California housing - estimation of the median house prices in the California area using California housing database. In this experiment, the learning rates are set to $0.01$ for all the algorithms. Fig. \ref{fig:error_california_housing} provides the normalized regression errors of the proposed algorithms, where it is observed that the DAT algorithm outperforms its competitors and can achieve a much higher nonlinear modeling power with respect to the rest of the algorithms.

Aside from the California housing data set, we also consider the regression of several benchmark real life and synthetic data from the corresponding data set repositories:
\begin{itemize}
  \item Kinematics \cite{delve} ($m=8$) - a realistic simulation of the forward dynamics of an $8$ link all-revolute robot arm. The task in all data sets is to predict the distance of the end-effector from a target. (among the existent variants of this data set, we used the variant with $m=8$, which is known to be highly nonlinear and medium noisy).
  \item Elevators \cite{keel} ($m=18$) - obtained from the task of controlling a F16 aircraft. In this case the goal variable is related to an action taken on the elevators of the aircraft.
  \item Pumadyn \cite{keel} ($m=32$) - a realistic simulation of the dynamics of Unimation Puma 560 robot arm. The task in the data set is to predict the angular acceleration of one of the robot arm's links.
  \item Bank \cite{ltorgo} ($m=32$) - generated from a simplistic simulator, which simulates the queues in a series of banks. Tasks are based on predicting the fraction of bank customers who leave the bank because of full queues (among the existent variants of this data set, we used the variant with $m=32$).
\end{itemize}

The learning rates of the LF, the VF, the FNF, the EMFNF, and the DAT algorithm are set to $\mu$, whereas it is set to $10\mu$ for the B-SAF and the CR-SAF algorithms, where $\mu=0.01$ for the kinematics, the elevators, and the bank data sets and $\mu=0.005$ for the pumadyn data set. In Table \ref{tab:error_table}, it is observed that the performance of the DAT algorithm is superior to its competitors since it achieves a much higher nonlinear modeling power with respect to the rest of the algorithms. Furthermore, the DAT algorithm achieves this superior performance with a computational complexity that is only linear in the regressor space dimensionality. Hence, the introduced algorithm can be used in real life big data problems.

\section{Concluding Remarks}\label{sec:Conclusion}
We study nonlinear regression of deterministic signals using trees,
where the space of regressors is partitioned using a nested tree
structure where separate regressors are assigned to each region. In
this framework, we introduce tree based regressors that both adapt
their regressors in each region as well as their tree structure to
best match to the underlying data while asymptotically achieving the
performance of the best linear combination of a doubly exponential
number of piecewise regressors represented on a tree. As shown in the
text, we achieve this performance with a computational complexity only
linear in the number of nodes of the tree. Furthermore, the introduced
algorithms do not require a priori information on the data such as
upper bounds or the length of the signal.  Since these algorithms
directly minimize the final regression error and avoid using any
artificial weighting coefficients, they readily outperform different
tree based regressors in our examples.  The introduced algorithms are
generic such that one can easily use different regressor or separation
functions or incorporate partitioning methods such as the RP trees in
their framework as explained in the paper.

\bibliographystyle{IEEEtran}
\bibliography{double_bib}

\begin{thebibliography}{10}
\providecommand{\url}[1]{#1}
\csname url@samestyle\endcsname
\providecommand{\newblock}{\relax}
\providecommand{\bibinfo}[2]{#2}
\providecommand{\BIBentrySTDinterwordspacing}{\spaceskip=0pt\relax}
\providecommand{\BIBentryALTinterwordstretchfactor}{4}
\providecommand{\BIBentryALTinterwordspacing}{\spaceskip=\fontdimen2\font plus
\BIBentryALTinterwordstretchfactor\fontdimen3\font minus
  \fontdimen4\font\relax}
\providecommand{\BIBforeignlanguage}[2]{{%
\expandafter\ifx\csname l@#1\endcsname\relax
\typeout{** WARNING: IEEEtran.bst: No hyphenation pattern has been}%
\typeout{** loaded for the language `#1'. Using the pattern for}%
\typeout{** the default language instead.}%
\else
\language=\csname l@#1\endcsname
\fi
#2}}
\providecommand{\BIBdecl}{\relax}
\BIBdecl

\bibitem{Singer3}
A.~C. Singer, G.~W. Wornell, and A.~V. Oppenheim, ``Nonlinear autoregressive
  modeling and estimation in the presence of noise,'' \emph{Digital Signal
  Processing}, vol.~4, no.~4, pp. 207--221, 1994.

\bibitem{Hero}
O.~J.~J. Michel, A.~O. Hero, and A.-E. Badel, ``Tree-structured nonlinear
  signal modeling and prediction,'' \emph{IEEE Transactions on Signal
  Processing}, vol.~47, no.~11, pp. 3027--3041, 1999.

\bibitem{drost}
R.~J. Drost and A.~C. Singer, ``Constrained complexity generalized context-tree
  algorithms,'' in \emph{IEEE/SP 14th Workshop on Statistical Signal
  Processing}, 2007, pp. 131--135.

\bibitem{CTW}
S.~S. Kozat, A.~C. Singer, and G.~C. Zeitler, ``Universal piecewise linear
  prediction via context trees,'' \emph{IEEE Transactions on Signal
  Processing}, vol.~55, no.~7, pp. 3730--3745, 2007.

\bibitem{volterra}
M.~Schetzen, \emph{The Volterra and Wiener Theories of Nonlinear
  Systems}.\hskip 1em plus 0.5em minus 0.4em\relax NJ: John Wiley \& Sons,
  1980.

\bibitem{saf}
M.~Scarpiniti, D.~Comminiello, R.~Parisi, and A.~Uncini, ``Nonlinear spline
  adaptive filtering,'' \emph{Signal Processing}, vol.~93, no.~4, pp. 772 --
  783, 2013.

\bibitem{fnf}
A.~Carini and G.~L. Sicuranza, ``Fourier nonlinear filters,'' \emph{Signal
  Processing}, vol.~94, no.~0, pp. 183 -- 194, 2014.

\bibitem{add1}
V.~Kekatos and G.~Giannakis, ``Sparse {V}olterra and polynomial regression
  models: Recoverability and estimation,'' \emph{IEEE Transactions on Signal
  Processing}, vol.~59, no.~12, pp. 5907--5920, 2011.

\bibitem{add2}
L.~Montefusco, D.~Lazzaro, and S.~Papi, ``Fast sparse image reconstruction
  using adaptive nonlinear filtering,'' \emph{IEEE Transactions on Image
  Processing}, vol.~20, no.~2, pp. 534--544, 2011.

\bibitem{add3}
Q.~Zhu, Z.~Zhang, Z.~Song, Y.~Xie, and L.~Wang, ``A novel nonlinear regression
  approach for efficient and accurate image matting,'' \emph{IEEE Signal
  Processing Letters}, vol.~20, no.~11, pp. 1078--1081, 2013.

\bibitem{add4}
R.~Mittelman and E.~Miller, ``Nonlinear filtering using a new proposal
  distribution and the improved fast {G}auss transform with tighter performance
  bounds,'' \emph{IEEE Transactions on Signal Processing}, vol.~56, no.~12, pp.
  5746--5757, 2008.

\bibitem{add5}
L.~Montefusco, D.~Lazzaro, and S.~Papi, ``Nonlinear filtering for sparse signal
  recovery from incomplete measurements,'' \emph{IEEE Transactions on Signal
  Processing}, vol.~57, no.~7, pp. 2494--2502, 2009.

\bibitem{add6}
W.~Zhang, B.-S. Chen, and C.-S. Tseng, ``Robust ${H}_\infty$ filtering for
  nonlinear stochastic systems,'' \emph{IEEE Transactions on Signal
  Processing}, vol.~53, no.~2, pp. 589--598, 2005.

\bibitem{add7}
H.~Zhao and J.~Zhang, ``A novel adaptive nonlinear filter-based pipelined
  feedforward second-order volterra architecture,'' \emph{IEEE Transactions on
  Signal Processing}, vol.~57, no.~1, pp. 237--246, 2009.

\bibitem{sp1}
L.~Ma, Z.~Wang, J.~Hu, Y.~Bo, and Z.~Guo, ``Robust variance-constrained
  filtering for a class of nonlinear stochastic systems with missing
  measurements,'' \emph{Signal Processing}, vol.~90, no.~6, pp. 2060 -- 2071,
  2010.

\bibitem{sp2}
W.~Yang, M.~Liu, and P.~Shi, ``${H}_\infty$ filtering for nonlinear stochastic
  systems with sensor saturation, quantization and random packet losses,''
  \emph{Signal Processing}, vol.~92, no.~6, pp. 1387 -- 1396, 2012.

\bibitem{sp3}
W.~Li and Y.~Jia, ``H-infinity filtering for a class of nonlinear discrete-time
  systems based on unscented transform,'' \emph{Signal Processing}, vol.~90,
  no.~12, pp. 3301 -- 3307, 2010.

\bibitem{sp4}
S.~Wen, Z.~Zeng, and T.~Huang, ``Reliable ${H}_\infty$ filtering for neutral
  systems with mixed delays and multiplicative noises,'' \emph{Signal
  Processing}, vol.~94, no.~0, pp. 23 -- 32, 2014.

\bibitem{sp5}
M.~F. Huber, ``Chebyshev polynomial kalman filter,'' \emph{Digital Signal
  Processing}, vol.~23, no.~5, pp. 1620 -- 1629, 2013.

\bibitem{Helmbold}
D.~P. Helmbold and R.~E. Schapire, ``Predicting nearly as well as the best
  pruning of a decision tree,'' \emph{Machine Learning}, vol.~27, no.~1, pp.
  51--68, 1997.

\bibitem{ml1}
O.-A. Maillard and R.~Munos, ``Linear regression with random projections,''
  \emph{Journal of Machine Learning Research}, vol.~13, pp. 2735 -- 2772, 2012.

\bibitem{ml2}
R.~Rosipal and L.~J. Trejo, ``Kernel partial least squares regression in
  {R}eproducing {K}ernel {H}ilbert {S}pace,'' \emph{Journal of Machine Learning
  Research}, vol.~2, pp. 97 -- 123, 2001.

\bibitem{ml3}
O.-A. Maillard and R.~Munos, ``Some greedy learning algorithms for sparce
  regression and classification with {M}ercer kernels,'' \emph{Journal of
  Machine Learning Research}, vol.~13, pp. 2735 -- 2772, 2012.

\bibitem{Singer1}
A.~C. Singer and M.~Feder, ``Universal linear prediction by model order
  weighting,'' \emph{IEEE Transactions on Signal Processing}, vol.~47, no.~10,
  pp. 2685--2699, 1999.

\bibitem{Moon1}
T.~Moon and T.~Weissman, ``Universal {FIR} {MMSE} filtering,'' \emph{IEEE
  Transactions on Signal Processing}, vol.~57, no.~3, pp. 1068--1083, 2009.

\bibitem{sayed_book}
A.~H. Sayed, \emph{Fundamentals of Adaptive Filtering}.\hskip 1em plus 0.5em
  minus 0.4em\relax NJ: John Wiley \& Sons, 2003.

\bibitem{Nascimento1}
V.~H. Nascimento and A.~H. Sayed, ``On the learning mechanism of adaptive
  filters,'' \emph{IEEE Transactions on Signal Processing}, vol.~48, no.~6, pp.
  1609--1625, 2000.

\bibitem{sayed2}
T.~Y. Al-Naffouri and A.~H. Sayed, ``Transient analysis of adaptive filters
  with error nonlinearities,'' \emph{IEEE Transactions on Signal Processing},
  vol.~51, no.~3, pp. 653--663, 2003.

\bibitem{Garcia1}
J.~Arenas-Garcia, A.~R. Figueiras-Vidal, and A.~H. Sayed, ``Mean-square
  performance of a convex combination of two adaptive filters,'' \emph{IEEE
  Transactions on Signal Processing}, vol.~54, no.~3, pp. 1078--1090, 2006.

\bibitem{RPTrees}
S.~Dasgupta and Y.~Freund, ``Random projection trees for vector quantization,''
  \emph{IEEE Transactions on Information Theory}, vol.~55, no.~7, pp.
  3229--3242, 2009.

\bibitem{CTW2}
Y.~Yilmaz and S.~S. Kozat, ``Competitive randomized nonlinear prediction under
  additive noise,'' \emph{IEEE Signal Processing Letters}, vol.~17, no.~4, pp.
  335--339, 2010.

\bibitem{ldf}
G.~David and A.~Averbuch, ``Hierarchical data organization, clustering and
  denoising via localized diffusion folders,'' \emph{Applied and Computational
  Harmonic Analysis}, vol.~33, no.~1, pp. 1 -- 23, 2012.

\bibitem{KDTrees}
J.~L. Bentley, ``Multidimensional binary search trees in database
  applications,'' \emph{IEEE Transactions on Software Engineering}, vol. SE-5,
  no.~4, pp. 333--340, 1979.

\bibitem{Takimoto1}
E.~Takimoto, A.~Maruoka, and V.~Vovk, ``Predicting nearly as well as the best
  pruning of a decision tree through dyanamic programming scheme,''
  \emph{Theoretical Computer Science}, vol. 261, pp. 179--209, 2001.

\bibitem{Takimoto2}
E.~Takimoto and M.~K. Warmuth, ``Predicting nearly as well as the best pruning
  of a planar decision graph,'' \emph{Theoretical Computer Science}, vol. 288,
  pp. 217--235, 2002.

\bibitem{double}
A.~V. Aho and N.~J.~A. Sloane, ``Some doubly exponential sequences,''
  \emph{Fibonacci Quarterly}, vol.~11, pp. 429--437, 1970.

\bibitem{Willems}
F.~M.~J. Willems, Y.~M. Shtarkov, and T.~J. Tjalkens, ``The context-tree
  weighting method: basic properties,'' \emph{IEEE Transactions on Information
  Theory}, vol.~41, no.~3, pp. 653--664, 1995.

\bibitem{Singer2}
A.~C. Singer, S.~S. Kozat, and M.~Feder, ``Universal linear least squares
  prediction: upper and lower bounds,'' \emph{IEEE Transactions on Information
  Theory}, vol.~48, no.~8, pp. 2354--2362, 2002.

\bibitem{linder1}
T.~Linder and G.~Lagosi, ``A zero-delay sequential scheme for lossy coding of
  individual sequences,'' \emph{IEEE Transactions on Information Theory},
  vol.~47, no.~6, pp. 2533--2538, 2001.

\bibitem{linder2}
A.~Gyorgy, T.~Linder, and G.~Lugosi, ``Efficient adaptive algorithms and
  minimax bounds for zero-delay lossy source coding,'' \emph{IEEE Transactions
  on Signal Processing}, vol.~52, no.~8, pp. 2337--2347, 2004.

\bibitem{Garcia2}
J.~Arenas-Garcia, V.~Gomez-Verdejo, and A.~R. Figueiras-Vidal, ``New algorithms
  for improved adaptive convex combination of {LMS} transversal filters,''
  \emph{IEEE Transactions on Instrumentation and Measurement}, vol.~54, no.~6,
  pp. 2239--2249, 2005.

\bibitem{logistic}
D.~W. Hosmer, S.~Lemeshow, and R.~X. Sturdivant, \emph{Applied Logistic
  Regression}.\hskip 1em plus 0.5em minus 0.4em\relax NJ: John Wiley \& Sons,
  2013.

\bibitem{Hazan}
E.~Hazan, A.~Agarwal, and S.~Kale, ``Logarithmic regret algorithms for online
  convex optimization,'' \emph{Machine Learning}, vol.~69, no. 2-3, pp.
  169--192, 2007.

\bibitem{Eweda}
E.~Eweda, ``Comparison of {RLS}, {LMS}, and sign algorithms for tracking
  randomly time-varying channels,'' \emph{IEEE Transactions on Signal
  Processing}, vol.~42, no.~11, pp. 2937--2944, 1994.

\bibitem{Garcia3}
J.~Arenas-Garcia, M.~Martinez-Ramon, V.~Gomez-Verdejo, and A.~R.
  Figueiras-Vidal, ``Multiple plant identifier via adaptive {LMS} convex
  combination,'' in \emph{2003 IEEE International Symposium on Intelligent
  Signal Processing}, 2003, pp. 137--142.

\bibitem{kozat}
S.~S. Kozat, A.~T. Erdogan, A.~C. Singer, and A.~H. Sayed, ``Steady state
  {M}{S}{E} performance analysis of mixture approaches to adaptive filtering,''
  \emph{IEEE Transactions on Signal Processing}, vol.~58, no.~8, pp.
  4050--4063, August 2010.

\bibitem{delve}
\BIBentryALTinterwordspacing
C.~E. Rasmussen, R.~M. Neal, G.~Hinton, D.~Camp, M.~Revow, Z.~Ghahramani,
  R.~Kustra, and R.~Tibshirani, ``Delve data sets.'' [Online]. Available:
  \url{http://www.cs.toronto.edu/~delve/data/datasets.html}
\BIBentrySTDinterwordspacing

\bibitem{keel}
J.~Alcala-Fdez, A.~Fernandez, J.~Luengo, J.~Derrac, S.~García, L.~Sánchez,
  and F.~Herrera, ``{KEEL} data-mining software tool: Data set repository,
  integration of algorithms and experimental analysis framework,''
  \emph{Journal of Multiple-Valued Logic and Soft Computing}, vol.~17, no. 2-3,
  pp. 255--287, 2011.

\bibitem{ltorgo}
\BIBentryALTinterwordspacing
L.~Torgo, ``Regression data sets.'' [Online]. Available:
  \url{http://www.dcc.fc.up.pt/~ltorgo/Regression/DataSets.html}
\BIBentrySTDinterwordspacing

\bibitem{lorenz}
E.~N. Lorenz, ``Deterministic nonperiodic flow,'' \emph{Journal of the
  Atmospheric Sciences}, vol.~20, no.~2, pp. 130--141, 1963.

\end{thebibliography}

\end{document}